\documentclass{article}

\usepackage[preprint]{neurips_2020}

\usepackage[utf8]{inputenc} 
\usepackage[T1]{fontenc}    
\usepackage{hyperref}       
\usepackage{url}            
\usepackage{booktabs}       
\usepackage{amsfonts}       
\usepackage{nicefrac}       
\usepackage{microtype}      

\usepackage{amsmath}
\usepackage{mathrsfs}
\usepackage{amssymb}
\usepackage{array}
\usepackage[font=small,labelfont=bf]{caption} 
\usepackage{subcaption}
\usepackage{float}
\usepackage{amsthm}
\usepackage{colortbl}
\usepackage{bm}

\usepackage{algorithm}
\usepackage[noend]{algorithmic}
\usepackage{makecell}

\usepackage{graphicx}
\usepackage{tabularx}
\usepackage{wrapfig}
\usepackage{enumitem}
\usepackage{xspace}
\usepackage[capitalise]{cleveref}

\theoremstyle{plain}
\newtheorem{thm}{Theorem}[section]

\newtheorem{prop}[thm]{Proposition}

\theoremstyle{definition}

\theoremstyle{remark}
\newtheorem*{rem}{Remark}

\newcommand{\E}{{\mathbb{E}}}
\newcommand{\V}{{\mathbb{V}}}
\newcommand{\C}{{\mathbb{C}}}
\newcommand{\R}{{\mathbb{R}}}
\newcommand{\N}{{\mathcal{N}}}

\newcommand{\I}{{\mathbb{I}}}
\newcommand{\U}{{\mathbb{U}}}

\newcommand{\xxi}{{\bm{x}_i}}
\newcommand{\param}{{\bm{\theta}}}
\newcommand{\lparam}{{\bm{\theta}}}

\newcommand{\Nk}{{N_k}}
\newcommand{\gi}{{\bm{g}_i}}
\newcommand{\gjk}{{\bm{g}_{j,k}}}
\newcommand{\gC}{{\bm{\hat{g}}}}
\newcommand{\Ai}{{a_i}}
\newcommand{\Aa}{{\bm{a}}}
\newcommand{\Ck}{{\bm{C}_k}}
\newcommand{\gb}{{\bm{g}_i}}  

\newcommand{\Di}{{I}}
\newcommand{\Do}{{O}}
\newcommand{\ck}{{\bm{c}_k}}
\newcommand{\dk}{{\bm{d}_k}}
\newcommand{\Ab}{{\bm{A}_i}}  
\newcommand{\Db}{{\bm{D}_i}}  
\newcommand{\AAA}{{\bm{A}}}
\newcommand{\DDD}{{\bm{D}}}

\newcommand{\SK}{{\sum_{k=1}^K}}
\newcommand{\SI}{{\sum_{i=1}^N}}
\newcommand{\SJ}{{\sum_{j=1}^\Nk}}
\newcommand{\SB}{{\sum_{i=1}^N}}  

\newcommand{\vv}[1]{{{\mathcal{\text{vec}}{{\{#1\}}}}}}

\DeclareMathOperator*{\argmin}{arg\,min}
\DeclareMathOperator*{\sign}{sign}

\newcommand{\dth}[1]{{\frac{\partial {#1}}{\partial\theta}}}

\renewcommand{\algorithmiccomment}[1]{\hfill$\triangleright$ #1}
\newcommand{\bA}{{$\mathcal{A}$} }
\newcommand{\bU}{{$\mathcal{U}$} }
\newcommand{\bAU}{{$\mathcal{AU}$} }

\newcommand{\Gluster}{\mbox{GC}\xspace}
\newcommand{\GC}{\mbox{GC}\xspace}
\newcommand{\SG}{\mbox{SG-B}\xspace}
\newcommand{\SGdB}{\mbox{SG-2B}\xspace}

\definecolor{mydarkblue}{rgb}{0,0.1,0.45}
\definecolor{amber}{rgb}{1.0, 0.75, 0.0}
\hypersetup{%
    colorlinks=true,
    linkcolor=mydarkblue,
    citecolor=mydarkblue,
    filecolor=mydarkblue,
    urlcolor=mydarkblue}
\newcommand{\threecolfigwidth}{0.32\textwidth}
\newcommand{\capshift}{-6mm}

\usepackage{etoolbox}
\newcommand{\zerodisplayskips}{%
  \setlength{\abovedisplayskip}{1pt}%
  \setlength{\belowdisplayskip}{1pt}%
  \setlength{\abovedisplayshortskip}{1pt}%
  \setlength{\belowdisplayshortskip}{1pt}}
\appto{\normalsize}{\zerodisplayskips}
\appto{\small}{\zerodisplayskips}
\appto{\footnotesize}{\zerodisplayskips}
\setlist[itemize]{leftmargin=20pt,itemsep=0.8pt}

\title{A Study of Gradient Variance in Deep Learning}

\author{%
    Fartash Faghri\hspace*{1cm}
    David Duvenaud\hspace*{1cm}
    David J.~Fleet\hspace*{1cm}
    Jimmy Ba\vspace*{0.2cm}\\
    University of Toronto, Vector Institute\\
    \texttt{\{faghri,duvenaud,fleet,jba\}@cs.toronto.edu}\\
}

\begin{document}

\maketitle

\begin{abstract}
    
The impact of gradient noise on  training  deep models is widely acknowledged 
but not well understood. In this context, we study the distribution of 
gradients during training.  We introduce a method, Gradient Clustering, to 
minimize the variance of average mini-batch gradient with stratified sampling.  
We prove that the variance of average mini-batch gradient is minimized if the 
elements are sampled from a weighted clustering in the gradient space.  We 
measure the gradient variance on common deep learning benchmarks and observe 
that, contrary to common assumptions, gradient variance increases during 
training, and smaller learning rates coincide with higher variance.
In addition, we introduce normalized gradient variance as a statistic that 
better correlates with the speed of convergence compared to gradient variance.

\end{abstract}

\section{Introduction}

Many machine learning tasks entail the minimization of the risk,
$\E_{\bm{x}}[\ell(\bm{x}; \param)]$,
where $\bm{x}$ is an i.i.d.\ sample from a data distribution, and $\ell$ is the 
per-example loss parametrized by $\param$.  In supervised learning, inputs and 
ground-truth labels comprise $\bm{x}$, and $\param$ is a vector of model 
parameters.
Empirical risk approximates the population risk by the risk of a sample set 
$\{\xxi\}_{i=1}^N$, the training set, as $L(\param)=\SI \ell(\xxi; \param)/N$.
Empirical risk is often minimized using gradient-based optimization 
(first-order methods).
For differentiable loss functions, the gradient of $\bm{x}$ is defined as 
$\dth{}\ell(\bm{x}; \param)$, i.e.\ the gradient of the loss with respect to 
the parameters evaluated at a point $\bm{x}$.
Popular in deep learning, Mini-batch Stochastic Gradient Descent (mini-batch 
SGD) iteratively takes small steps in the opposite direction of the average 
gradient of $B$ training samples.
The mini-batch size is a hyperparameter that provides flexibility in trading 
per-step computation time for potentially fewer total steps. In GD the 
mini-batch is the entire training set while in SGD it is a single sample.

In general, using any unbiased stochastic estimate of the gradient and 
sufficiently small step sizes, SGD is guaranteed to converge to a minimum for 
various function classes~\citep{robbins1951stochastic}. Common convergence 
bounds in stochastic optimization improve with smaller gradient 
variance~\citep{bottou2018optimization}. Mini-batch SGD is said to converge 
faster because the variance of the gradient estimates is reduced by a rate 
linear in the mini-batch size. In practice however, we observe {\bf diminishing 
returns}\/ in speeding up the training of almost any deep model on deep 
learning benchmarks~\citep{shallue2018measuring}.
One explanation not studied by previous work is that the variance numerically 
reaches zero. The transition point to diminishing returns is known to depend on 
the choice of data, model and optimization method.
\citet{zhang2019algorithmic} observed that the limitation of acceleration in 
large batches is reduced when momentum or preconditioning is used.  Other works 
suggest that very small mini-batch sizes can still converge fast enough using 
a collection of tricks~\citep{golmant2018, masters2018revisiting, lin2018}. One 
hypothesis is that the stochasticity due to small mini-batches improves 
generalization by finding ``flat minima'' and avoiding ``sharp 
minima''~\citep{goodfellow2015qualitatively,keskar2017large}.  But this 
hypothesis does not explain why diminishing returns also happens in the 
training loss.

Motivated by the diminishing returns phenomena, we study and model the 
distribution of the gradients.
In the noisy gradient view, the average mini-batch gradient (or the mini-batch 
gradient) is treated as an unbiased estimator of the expected gradient where 
increasing the mini-batch size reduces the variance of this estimator.
We propose a distributional view and argue that knowledge of the gradient 
distribution can be exploited to analyze and improve optimization speed as well 
as generalization to test data. A mean-aware optimization method is at best as 
strong as a distributional-aware optimization method.
In our distributional view, the mini-batch gradient is only an estimate of the 
mean of the gradient distribution.

{\bf Questions:} We identify the following understudied questions about the 
gradient distribution.
\begin{itemize}
    \item[] {\bf Structure of gradient distribution.}\/ Is there structure in 
        the distribution over gradients of standard learning problems?
    \item[] {\bf Impact of gradient distribution on optimization.}\/ What 
        characteristics of the gradient distribution correlate with the 
        convergence speed and the minimum training/test loss reached?
    \item[] {\bf Impact of optimization on gradient distribution.}\/ To what 
        extent do the following factors affect the gradient distribution:  data 
        distribution, learning rate, model architecture, mini-batch size, 
        optimization method, and the distance to local optima?
\end{itemize}

\begin{figure}[t]
    \centering
    \begin{subfigure}[b]{\threecolfigwidth}
        \includegraphics[width=\textwidth]{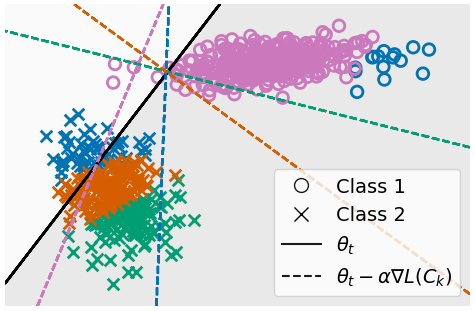}
        \vspace*{\capshift}
        \caption{GD step 0}
    \end{subfigure}
    \begin{subfigure}[b]{\threecolfigwidth}
        \includegraphics[width=\textwidth]{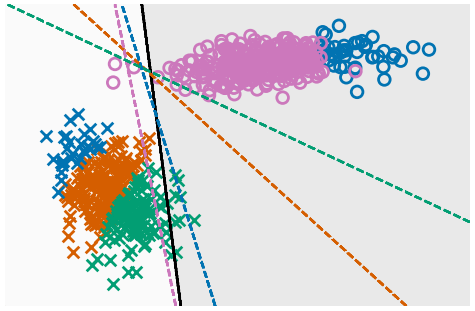}
        \vspace*{\capshift}
        \caption{GD step 1}
    \end{subfigure}
    \begin{subfigure}[b]{\threecolfigwidth}
        \includegraphics[width=\textwidth]{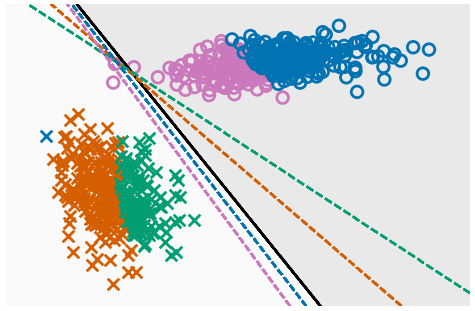}
        \vspace*{\capshift}
        \caption{GD step 2}
    \end{subfigure}
    \caption{{\bf Example of clusters found using Gradient Clustering.}\/ 
    A linear classifier visualized during training with gradient descent on 
    2 linearly separable classes (o, x).  Gradients are assigned to $4$ 
    clusters (different colors) using Gradient Clustering (\Gluster).  Black 
    line depicts current decision boundary.  Colored dashed lines depict 
    decision boundaries predicted from current boundary and each of the $4$ 
    individual clusters.  Here, blue points belong to both classes; they have 
    similar gradients, but are far apart in input space. By exploiting the 
    knowledge of \Gluster we can get low variance average mini-batch gradients.
    }
    \label{fig:2dvis_single}
    \vspace*{-14pt}
\end{figure}

{\bf Contributions:}
\begin{itemize}
    \item[] {\bf Exploiting clustered distributions.}\/ We consider gradient 
        distributions with distinct modes, i.e.\ the gradients can be 
        clustered.
        We prove that the variance of average mini-batch gradient is minimized 
        if the elements are sampled from a weighted clustering in gradient 
        space (\cref{sec:gvar}).
    \item[] {\bf Efficient clustering to minimize variance.}\/ We propose 
        Gradient Clustering (\Gluster) as a computationally efficient method 
        for clustering in the gradient space (\cref{sec:gluster}).
        \cref{fig:2dvis_single} shows an example of clusters found by \Gluster.
    \item[] {\bf Relation between gradient variance and optimization.}\/ We 
        study the gradient variance on common deep learning benchmarks (MNIST, 
        CIFAR-10, and ImageNet) as well as Random Features models recently 
        studied in deep learning theory (\cref{sec:exp}). We observe that 
        gradient variance increases during training, and smaller learning rates 
        coincide with higher variance.
    \item[] {\bf An alternative statistic.}\/ We introduce normalized gradient 
        variance as a statistic that better correlates with the speed of 
        convergence compared to gradient variance (\cref{sec:exp}).
\end{itemize}

We emphasize that some of our contributions are primarily empirical yet 
unexpected. We encourage the reader to predict the behaviour of gradient 
variance before reaching our experiments section. We believe our results 
provide an opportunity for future theoretical and empirical work.

\section{Related Work}

{\bf Modeling gradient distribution.}\/
Despite various assumptions on the mini-batch gradient variance, there are 
limited studies of this statistic during the training of deep learning models.
It is common to assume bounded variance in convergence 
analyses~\citep{bottou2018optimization}. Works on variance reduction propose 
alternative estimates of the gradient mean with low 
variance~\citep{leroux2012stochastic, johnson2013accelerating} but they do not 
plot the variance which is the actual quantity they seek to reduce.  Their 
ineffectiveness in deep learning has been observed but still requires 
explanation~\citep{defazio2019ineffectiveness}.
There are a few works that present gradient variance 
plots~\citep{mohamed2019monte, wen2019interplay} but they are usually for 
a single gradient coordinate and synthetic problems.
The Central limit theorem is also used to argue that the distribution of the 
mini-batch gradient is a Gaussian~\citep{zhu2018anisotropic}, which has been 
challenged only recently~\citep{simsekli2019tail}. There also exists a link 
between the Fisher~\citep{amari1998natural}, Neural Tangent 
Kernel~\citep{jacot2018neural}, and the gradient covariance 
matrix~\citep{martens2014new, kunstner2019limitations, thomas2020interplay}. As 
such, any analysis of one~\citep[e.g.][]{karakida2019pathological} could 
potentially be used to understand others.

{\bf The variance is rarely used for improving optimization.}
\citet{le2011improving} considered the difference between the covariance matrix 
of the gradients and the Fisher matrix and proposed incorporating the 
covariance matrix as a measure of model uncertainty in optimization. It has 
also been suggested that the division by the second moments of the gradient in 
Adam can be interpreted as variance adaptation~\citep{kunstner2019limitations}.
There are myriad papers on ad-hoc sampling and re-weighting methods for 
reducing dataset imbalance and increasing data 
diversity~\citep{bengio2008adaptive, jiang2017mentornet, vodrahalli2018, 
jiang2019accelerating}.  Although we do not use Gradient Clustering for 
optimization, the formulation can be interpreted as a unifying approach that 
defines variance reduction as an objective.

{\bf Clustering gradients.}\/
Methods related to gradient clustering have been proposed in low-variance 
gradient estimation~\citep{hofmann2015variance, zhao14accelerating, 
chen2019fast} supported by promising theory.  However, these methods have 
either limited their experiments to linear models or treated a deep model as 
a linear one.  Our proposed \GC method performs efficient clustering in the 
gradient space with very few assumptions.
\GC is also related to works on model visualization where the entire training 
set is used to understand the behaviour of a model~\citep{raghu2017svcca}.

\section{Mini-batch Gradient with Stratified Sampling}\label{sec:gvar}
An important factor affecting optimization trade-offs is the diversity of 
training data. SGD entails a sampling process, often uniformly sampling from 
the training set.  However, as illustrated in the following example, uniform 
sampling is not always ideal. Suppose there are duplicate data points in 
a training set. We can save computation time by removing all but one of the 
duplicates.  To get the same gradient mean in expectation, it is sufficient to 
rescale the gradient of the remaining sample in proportion to the number of 
duplicates. In this example, mini-batch SGD will be inefficient because 
duplicates increase the variance of the average gradient mean.

Suppose we are given i.i.d.\ training data, $\{\xxi\}_{i=1}^N$, and 
a partitioning of their gradients, $\gi=\dth{}\ell(\xxi; \param)$, into $K$ 
subsets, where $\Nk$ is the size of the $k$-th set. We can estimate the 
gradient mean on the training set, $\bm{g}=\dth{}L(\param)$, by averaging $K$ 
gradients, one from each of $K$ subsets, uniformly sampled:
\begin{align}
    \gC(\Aa) &= \frac{1}{N} \SK \Nk\, \gjk\,,\quad j \sim \U[1, \Nk]\,,
    \label{eq:strat_sampl}
\end{align}
where $g_{.,k} = \{\gi|\Ai=k\}$ are the gradients from a subset $k$, 
$\gjk\in\R^d$ is the gradient of the $j$-th sample in the $k$-th subset, and 
$\Aa\in\{1,\ldots,K\}^N$ where $\Ai$ is the index of the cluster to which 
$i$-th data point is assigned, so $\Nk=\SI \I(\Ai=k)$.~\footnote{While $j,k$ 
index data based on the partitioning, $i$ indexes the training points 
independent of partitioning.}  Each sample is treated as a representative of 
its subset and weighted by the size of that subset.  In the limit of $K=N$, we 
recover the batch gradient mean used in GD and for $K=1$ we recover the 
single-sample stochastic gradient in SGD.

\begin{prop}\label{thm:gvar}
    {\bf (Bias/Variance of Mini-batch Gradient with Stratified Sampling).}\/
    For any partitioning of data, the estimator of the gradient mean using 
    stratified sampling (\cref{eq:strat_sampl}) is unbiased ($\E[\gC]=g$) and
    $\V[\gC] = N^{-2} \SK \Nk^2 \V[\gjk]$,
    where $\V[\cdot]$ is defined as the trace of the covariance matrix.
    (Proof in \cref{sec:gvar_proof})
\end{prop}

\begin{rem}
    In a dataset with duplicate samples, the gradients of duplicates do not 
    contribute to the variance if assigned to the same partition with no other 
    data points.
\end{rem}

\subsection{Weighted Gradient Clustering}
\label{sec:cluster}

Suppose, for a given number of clusters, $K$, we want to find the 
optimal partitioning, i.e., one that minimizes the variance of the gradient 
mean estimator, $\gC$.  For $d$-dimensional gradient vectors, minimizing the 
variance in \cref{thm:gvar}, is equivalent to finding a weighted clustering of 
the gradients of data points,
\begin{align}
    \min_\Aa \V[\gC(\Ai)]
    &= \min_\Aa \SK \Nk^2 \V[\gjk]
    = \min_{\bm{C}, \Aa} \SK \SI \Nk\, \|\Ck-\gi\|^2 \,\I(\Ai=k)\,,
    \label{eq:obj}
\end{align}
where a cluster center, $\Ck\in\R^d$, is the average of the gradients in the 
$k$-th cluster, and ${\V[\gjk]=\frac{1}{\Nk} \SI \|\Ck-\gi\|^2 \,\I(\Ai=k)}$.  
If we did not have the factor $\Nk$, this objective would be equivalent to the 
K-Means objective.  The additional $\Nk$ factors encourage larger clusters to 
have lower variance, with smaller clusters comprising scattered data points.  

If we could store the gradients for the entire training set, the clustering 
could be performed iteratively as a form of block coordinate descent, 
alternating between the following {\em Assignment}\ and {\em Update}\ steps,
i.e., computing the cluster assignments and then the cluster centers:
\noindent\begin{tabular}{>{\centering\arraybackslash} m{.45\textwidth}
    >{\centering\arraybackslash} m{.45\textwidth}}
    \setlength{\tabcolsep}{10pt}
    \centering
 \vspace*{-0.3cm}
    \begin{equation}
    {\text{\bA:}}\quad \Ai = \argmin_k \Nk\, \|\Ck-\gi\|^2
    \label{Estep}
    \end{equation}
    &
 \vspace*{-0.3cm}
    \begin{equation}
    {\text{\bU:}}\quad \Ck = \frac{1}{\Nk} \SI \gi\, \I(\Ai=k) 
    \label{Mstep}
    \end{equation}
\end{tabular}
 \vspace*{-0.3cm}

The \bA step is still too complex given the $\Nk$ multiplier. As such, we first 
solve it for fixed cluster sizes then update $\Nk$ before another \bU step.
These updates are similar to Lloyd's algorithm for K-Means, but 
with the $\Nk$ multipliers, and to Expectation-Maximization for Gaussian
Mixture Models, but here we use hard assignments.  In contrast, the additional 
$\Nk$ multiplier makes the objective more complex in that performing \bAU 
updates does not always guarantee a decrease in the clustering objective.

\subsection{Efficient Gradient Clustering (\Gluster)}\label{sec:gluster}

Performing exact \bAU updates (\cref{Estep,Mstep}) is computationally expensive 
as they require the gradient of every data point. Deep learning libraries 
usually provide efficient methods that compute average mini-batch gradients 
without ever computing full individual gradients.
We introduce Gradient Clustering (\Gluster) for performing efficient \bAU 
updates by breaking them into per-layer operations and introducing a low-rank 
approximation to cluster centers.

\begin{figure}[t]
\begin{minipage}[t]{0.33\textwidth}
    \centering
    \begin{algorithm}[H]
        \centering
        \caption{\bA step using \cref{eq:fc:Estep}}
        \begin{algorithmic}
            \FOR{$i=1$ {\bf to} $N$}
              \FOR{$k=1$ {\bf to} $K$} 
                \FOR{$l=1$ {\bf to} $L$} 
                  \STATE $D_{kl} = \| \bm{C}_{kl} - \bm{g}_{il}\|^2$
                \ENDFOR
              \ENDFOR
              \STATE $\bm{S} = \sum_l D_{\cdot l}$
              \STATE $\Ai = \argmin_k \Nk \bm{S}$
            \ENDFOR
        \end{algorithmic}
        \label{alg:A_step}
    \end{algorithm}
\end{minipage}
\hfill
\begin{minipage}[t]{0.30\textwidth}
    \centering
    \begin{algorithm}[H]
        \centering
        \caption{$\Nk$ update}
        \begin{algorithmic}
            \STATE $\Nk = 0,\quad \forall k={1,\cdots,K}$
            \FOR{$i=1$ {\bf to} $N$}
              \STATE $N_{\Ai} \mathrel{+}= 1$
            \ENDFOR
        \end{algorithmic}
        \label{alg:A_step}
    \end{algorithm}
\end{minipage}
\hfill
\begin{minipage}[t]{0.33\textwidth}
    \centering
    \begin{algorithm}[H]
        \centering
        \caption{\bU step using \cref{eq:fc:Mstep}}
        \begin{algorithmic}
            \STATE $\bm{C}_k = 0,\quad \forall k={1,\cdots,K}$
            \FOR{$i=1$ {\bf to} $N$}
              \FOR{$l=1$ {\bf to} $L$} 
                \STATE $\bm{C}_{\Ai,l} \mathrel{+}= \bm{g}_{il} / N_{\Ai}$
              \ENDFOR
            \ENDFOR
        \end{algorithmic}
        \label{alg:U_step}
    \end{algorithm}
\end{minipage}
    \caption{Steps in Gradient Clustering}
    \label{alg:full}
    \vspace*{-14pt}
\end{figure}

For any feed-forward network, we can decompose terms in \bAU updates into 
independent per-layer operations as shown in \cref{alg:full}. The main 
operations are computing $\|\bm{C}_{kl} - \bm{g}_{il}\|^2$ and cluster updates 
$\bm{C}_{\Ai,l} \mathrel{+}= \bm{g}_{il} / N_{\Ai}$ per layer $l$; henceforth, 
we drop the layer index for simplicity.

For a single fully-connected layer, we denote the layer weights by $\lparam\in 
\R^{\Di\times \Do}$, where $\Di$ and $\Do$ denote the input and output 
dimensions for the layer.  We denote the gradient with respect to $\lparam$ for 
the training set by $\bm{g}=\AAA \DDD^\top$, where $\AAA\in \R^{\Di\times N}$ 
comprises the input activations to the layer, and $\DDD\in \R^{\Do\times N}$ 
represents the gradients with respect to the layer outputs.
The coordinates of cluster centers corresponding to this layer are denoted by 
$C\in \R^{K\times \Di\times \Do}$.  We index the clusters using $k$ and the 
data by $i$. The $k$-th cluster center is approximated as
${\Ck=\ck \dk^\top}$, using vectors ${\ck\in \R^\Di}$ and ${\dk\in\R^\Do}$.

In the \bA step we need to compute ${\|\Ck-\gb\|_F^2}$ as part of the 
assignment cost, where $\|\cdot\|_F$ is the Frobenius-norm. We expand this term 
into three inner-products, and compute them separately. In particular, the term 
$\vv{\Ck}\odot\vv{\gb}$ can be written as,
\begin{align}
    \vv{\Ck}\odot \vv{\Ab \Db^\top}
    = (\Ab \odot \ck)(\Db\odot \dk)\,,
    \label{eq:fc:Estep}
\end{align}
where $\odot$ denotes inner product, and the RHS is the product of two scalars.  
Similarly, we compute the other two terms in the expansion of the assignment 
cost, i.e.\ $\vv{\Ck}\odot\vv{\Ck}$ and $\vv{\gb}\odot\vv{\gb}$ 
(\citet{goodfellow2015efficient} proposed a similar idea to compute the 
gradient norm).

The \bU step in \cref{Mstep} is written as,
$\ck \dk^\top
= \Nk^{-1} \SB \Ab \Db^\top \I(\Ai=k)$.
This equation might have no exact solution for $\ck$ and $\dk$ because the sum 
of rank-$1$ matrices is not necessarily rank-$1$.  One approximation is the 
min-Frobenius-norm solution to $\ck, \dk$ using truncated SVD, where we use 
left and right singular-vectors corresponding to the largest singular-value of 
the RHS.
However, the following updates are exact if
activations and gradients of the outputs are uncorrelated, i.e.\ $\E_i[\Ab 
\Db]=\E_i[\Ab]E_i[\Db]$ (similar to assumptions in 
K-FAC~\citep{martens2015kfac}),
\begin{align}
    \ck = \frac{1}{\Nk} \SB \Ab\I(\Ai=k)
    \qquad\qquad
    \dk = \frac{1}{\Nk} \SB \Db\I(\Ai=k)\,.
    \label{eq:fc:Mstep}
\end{align}
\vspace*{-0.3cm}

In \cref{sec:conv}, we describe similar update rules for convolutional layers 
and in \cref{sec:comp}, we provide complexity analysis of \Gluster.
We can make the cost of \Gluster negligible by making sparse incremental 
updates to cluster centers using mini-batch updates. The assignment step can 
also be made more efficient by processing only a portion of data as is common 
for training on large datasets. The rank-$1$ approximation can be extended to 
higher rank approximations with multiple independent cluster centers though 
with challenges in the implementation.

\section{Experiments}\label{sec:exp}

In this section, we evaluate the accuracy of estimators of the gradient mean.  
This is a surrogate task for evaluating the performance of a model of the 
gradient distribution. We compare our proposed \GC estimator to average 
mini-batch Stochastic Gradient (\SG), and \SG with double the mini-batch size 
(\SGdB).
\SGdB is an important baseline for two reasons. First, it is a competitive 
baseline that always reduces the variance by a factor of $2$ and requires at 
most twice the memory size and twice the run-time per 
mini-batch~\citep{shallue2018measuring}.
Second, the extra overhead of \Gluster is approximately the same as keeping an 
extra mini-batch in the memory when the number of clusters is equal to the 
mini-batch size.
We also include Stochastic Variance Reduced Gradient 
(SVRG)~\citep{johnson2013accelerating} as a method with the sole objective of 
estimating gradient mean with low variance.

We compare methods on \textbf{a single trajectory of mini-batch SGD} to 
decouple the optimization from gradient estimation.  That is, we do not train 
with any of the estimators (hence no `D' in \SG and \SGdB).  This allows us to 
continue analyzing a method even after it fails in reducing the variance.  For 
training results using \SG, \SGdB and, SVRG, we refer the reader to 
\citet{shallue2018measuring, defazio2019ineffectiveness}.  For training with 
\GC, it suffices to say that behaviours observed in this section are directly 
related to the performance of \GC used for optimization.

As all estimators in this work are unbiased, the estimator with lowest variance 
is better estimating the gradient mean.
We define {\it Average Variance}\/ (variance in short) as the average over all 
coordinates of the variance of the gradient mean estimate for a fixed model 
snapshot.  Average variance is the normalized trace of the covariance matrix 
and of particular interest in random matrix theory~\citep{tao2012topics}.  We 
also measure {\it Normalized Variance}, defined as $\V[g]/E[g^2]$ where the 
variance of a $1$-dimensional random variable is divided by its second 
non-central moment.  In signal processing, the inverse of this quantity is the 
signal to noise ratio (SNR). If SNR is less than one (normalized gradient 
larger than one), the power of the noise is greater than the signal. Additional 
details of the experimental setup can be found in \cref{app:exp}.

\subsection{MNIST: Low Variance, CIFAR-10: Noisy Estimates, ImageNet: No 
Structure}\label{sec:exp_image}

\begin{figure}[t]
    \centering
    \begin{subfigure}[b]{\threecolfigwidth}
        \includegraphics[width=\textwidth]{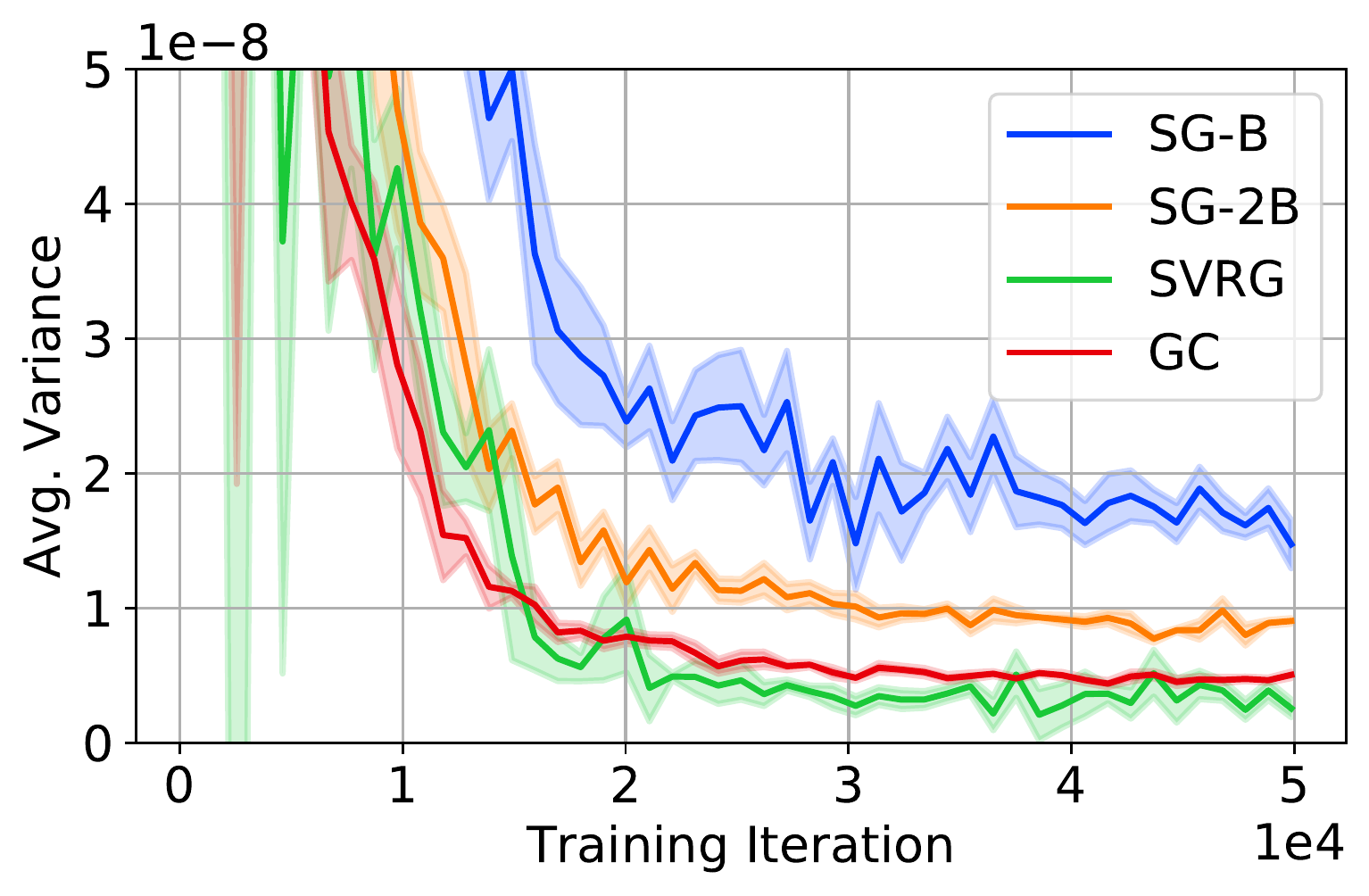}
        \vspace*{\capshift}
        \caption{MLP on MNIST}
        \label{fig:mnist_var}
    \end{subfigure}
    \begin{subfigure}[b]{\threecolfigwidth}
        \includegraphics[width=\textwidth]{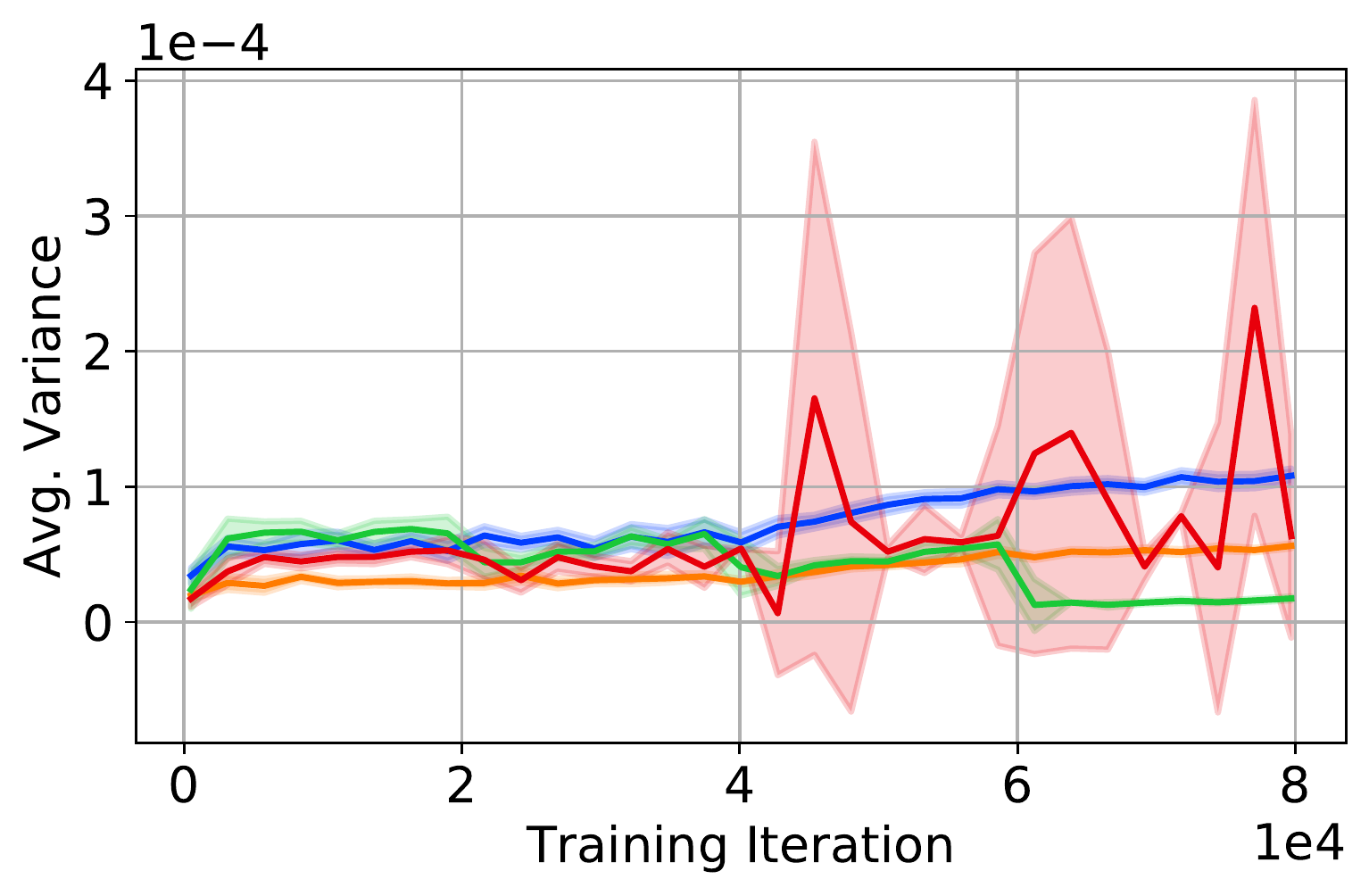}
        \vspace*{\capshift}
        \caption{ResNet8 on CIFAR-10}
        \label{fig:cifar10_var}
    \end{subfigure}
    \begin{subfigure}[b]{\threecolfigwidth}
        \includegraphics[width=\textwidth]{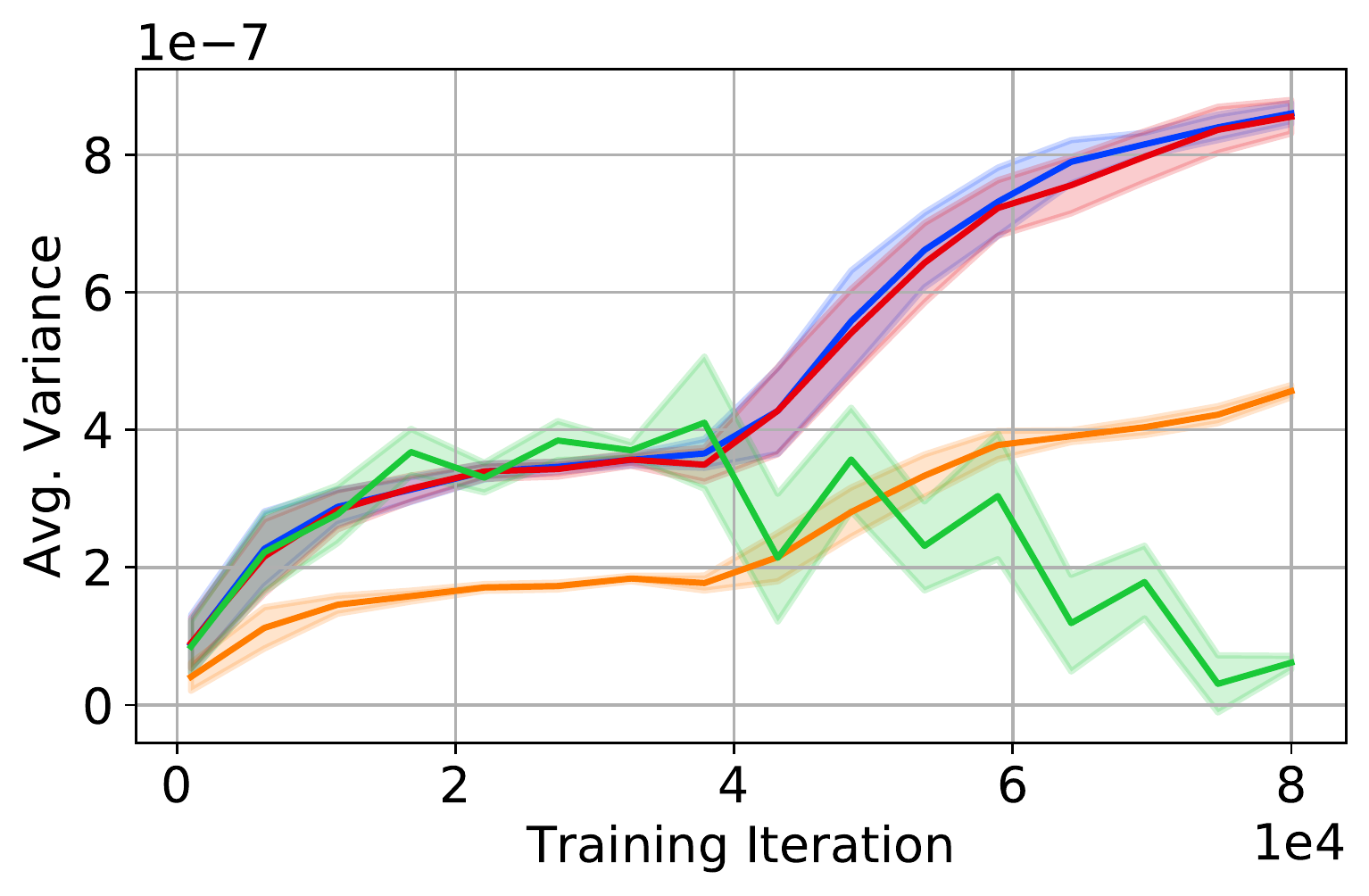}
        \vspace*{\capshift}
        \caption{ResNet18 on ImageNet}
        \label{fig:imagenet_var}
    \end{subfigure}\\
    \begin{subfigure}[b]{\threecolfigwidth}
        \includegraphics[width=\textwidth]{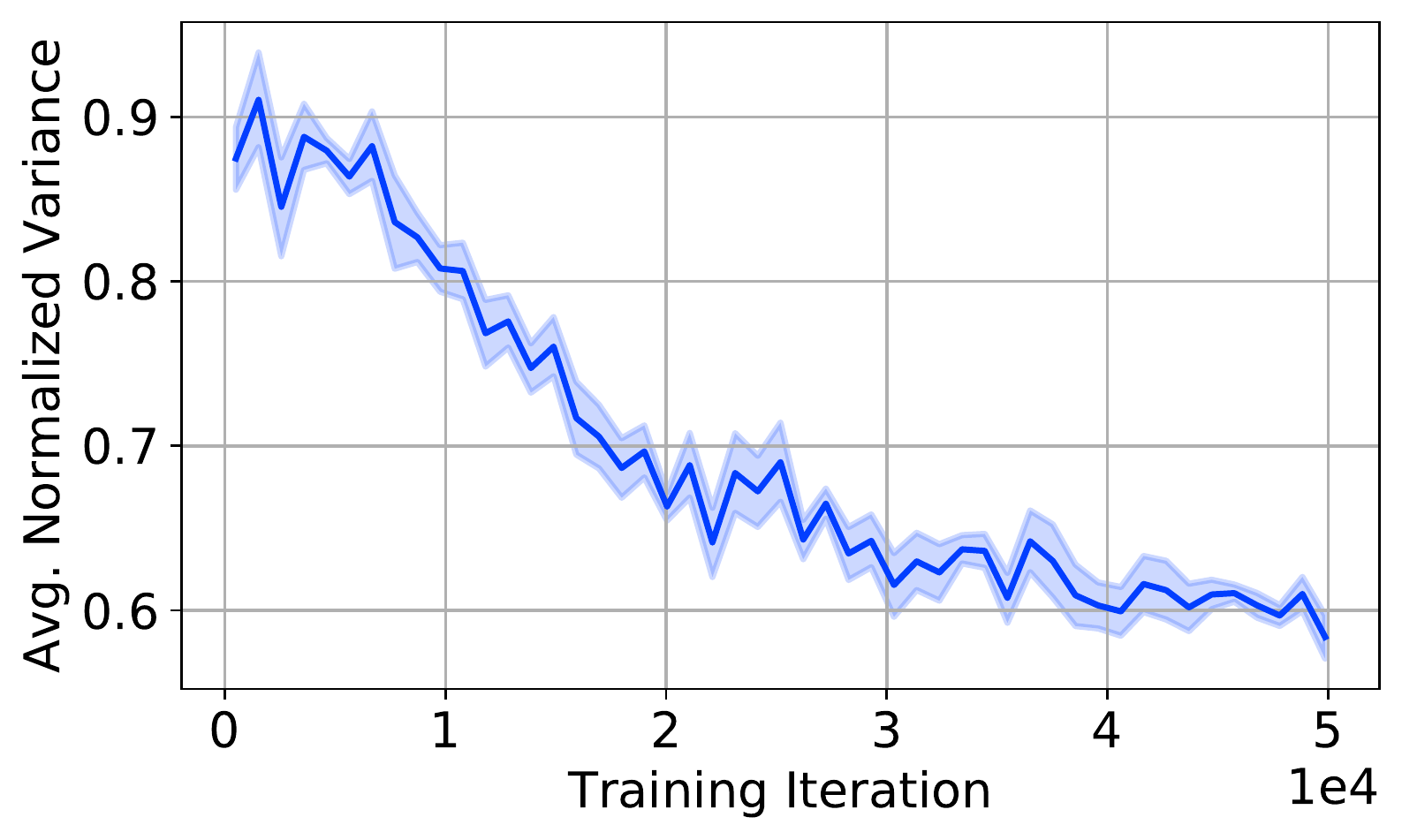}
        \vspace*{\capshift}
        \caption{MLP on MNIST}
        \label{fig:mnist_nvar}
    \end{subfigure}
    \begin{subfigure}[b]{\threecolfigwidth}
        \includegraphics[width=\textwidth]{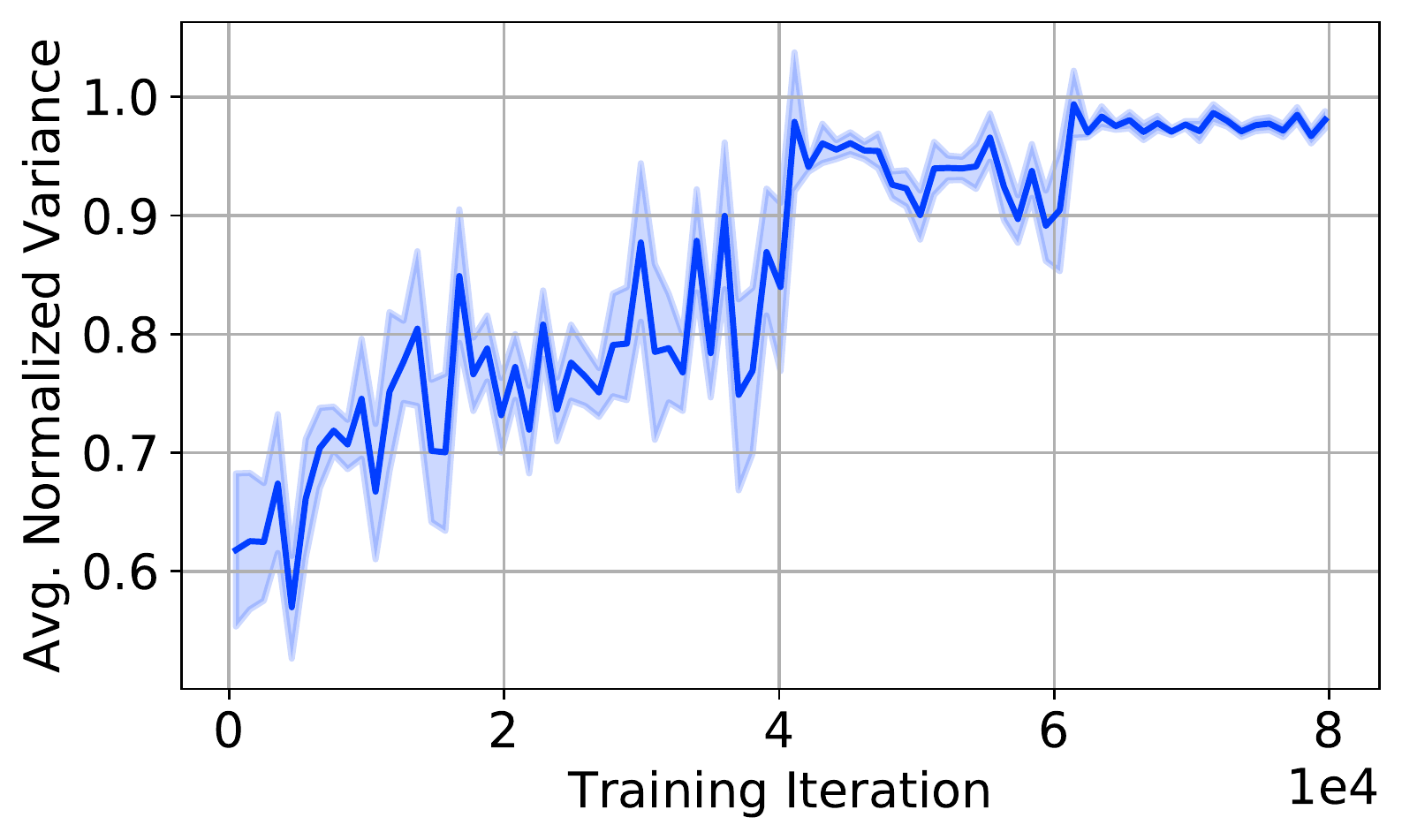}
        \vspace*{\capshift}
        \caption{ResNet8 on CIFAR-10}
        \label{fig:cifar10_nvar}
    \end{subfigure}
    \begin{subfigure}[b]{\threecolfigwidth}
        \includegraphics[width=\textwidth]{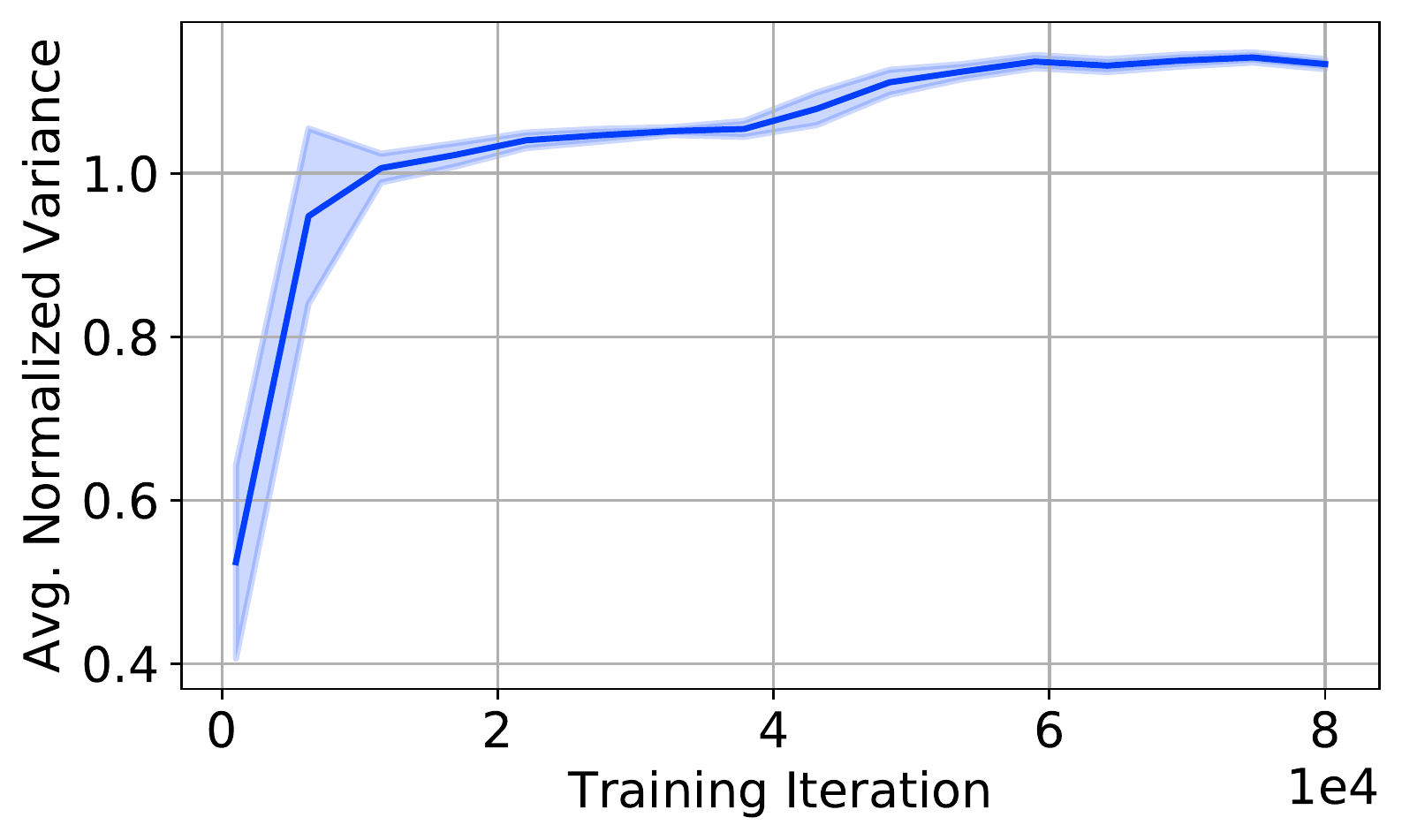}
        \vspace*{\capshift}
        \caption{ResNet18 on ImageNet}
        \label{fig:imagenet_nvar}
    \end{subfigure}
    \caption{{\bf Image classification models.}\/ Variance (top) and normalized 
    variance plots (bottom).
    We observe:
    normalized variance correlates with optimization difficulty,
    variance is decreasing on MNIST but increasing on CIFAR-10 and ImageNet,
    and variance fluctuates with \GC on CIFAR-10.
    }
    \label{fig:image}
    \vspace*{-14pt}
\end{figure}

In this section, we study the evolution of gradient variance during training of an 
MLP on MNIST~\citep{lecun1998gradient}, ResNet8~\citep{he2016deep} on 
CIFAR-10~\citep{krizhevsky2009learning}, and ResNet18 on 
ImageNet~\citep{deng2009imagenet}. Curves shown are from a single run and 
statistics are smoothed out over a rolling window.  The standard deviation 
within the window is shown as a shaded area.

\textbf{Normalized variance correlates with the time required to improve 
accuracy.}
In \cref{fig:mnist_var,fig:cifar10_var,fig:imagenet_var}, the variance of \SGdB 
is always half the variance of {\SG}. A drawback of the variance is that it is 
not comparable across different problems. For example, on CIFAR-10 the variance 
of all methods reaches $10^{-4}$ while on ImageNet where usually $10\times$ 
more iterations are needed, the variance is below $10^{-6}$. In 
\cref{fig:mnist_nvar,fig:cifar10_nvar,fig:imagenet_nvar}, normalized variance 
better correlates with the convergence speed.  Normalized variance on both 
MNIST and CIFAR-10 is always below $1$ while on ImageNet it quickly goes above 
$1$ (noise stronger than gradient).
Notice that the denominator in the normalized variance is shared between all 
methods on the same trajectory of mini-batch SGD\@. As such, the normalized 
variance retains the relation of curves and is a scaled version of variance 
where the scaling varies during training as the norm of the gradient changes.  
For clarity, we only show the curve for {\SG}.

{\bf How does the difficulty of optimization change during training?}\/
The variance on MNIST for all methods is constantly decreasing 
(\cref{fig:mnist_var}), i.e.\ the strength of noise decreases as we get closer 
to a local optima. These plots suggest that training an MLP on MNIST satisfies 
the Strong Growth Condition (SGC)~\citep{schmidt2013fast} as the variance is 
numerically zero (below $10^{-8}$).
Normalized variance (\cref{fig:mnist_nvar}) decreases over time and is well 
below $1$ (gradient mean has larger magnitude than the variance). SVRG performs 
particularly well by the end of the training because the training loss has 
converged to near zero (cross-entropy less than $0.005$).  Promising published 
results with SVRG are usually on datasets similar to MNIST where the loss 
reaches relatively small values.
In contrast, on both CIFAR-10 (\cref{fig:cifar10_var,fig:cifar10_nvar}) and 
ImageNet (\cref{fig:imagenet_var,fig:imagenet_nvar}), the variance and 
normalized variance of all methods increase during training and especially 
after the learning rate drops.  This means gradient variance depends on the 
distance to local optima.  We hypothesize that the gradient of each training 
point becomes more unique as training progresses.

\textbf{Variance can widely change during training but it happens only on 
a particularly noisy data.}
On CIFAR-10, the variance of \Gluster suddenly goes up but comes back down 
before any updates to the cluster centers (\cref{fig:cifar10_var}) while the 
variance of SVRG monotonically increases between updates.  To explain these 
behaviours, notice that immediately after cluster updates, \Gluster and SVRG 
should always have at most the same average variance as {\SG}.   We observed 
this behaviour consistently across different architectures such as other 
variations of ResNet and {VGG} on CIFAR-10.  \cref{fig:cifar10_noise} shows the 
effect of adding noise on CIFAR-10.  Label 
smoothing~\citep{szegedy2016rethinking} reduces fluctuations but not 
completely. On the other hand, label corruption, where we randomly change the 
labels for $10\%$ of the training data eliminates the fluctuations. We 
hypothesize that the model is oscillating between different states with 
significantly different gradient distributions.  The experiments with corrupt 
labels suggest that mislabeled data might be the cause of fluctuations such 
that having more randomness in the labels forces the model to ignore originally 
mislabeled data.

\textbf{Is the gradient distribution clustered in any dataset?}
The variance of \GC on MNIST (\cref{fig:mnist_var}) is consistently lower than 
\SGdB which means it is exploiting clustering in the gradient space. On 
CIFAR-10 (\cref{fig:cifar10_var}) the variance of \GC is lower than \SG but not 
lower than \SGdB except when fluctuating. The improved variance is more 
noticeable when training with corrupt labels.
On ImageNet (\cref{fig:imagenet_var,fig:imagenet_nvar}), the variance of \GC is 
overlapping with {\SG}. An example of a gradient distribution where \GC is 
overlapping with \SG is a uniform distribution. Gradient distribution can still 
be structured.  For example, there could be clusters in subspaces of the 
gradient space.

\subsection{Random Features Models: How Does Overparametrization Affect the 
Variance?}\label{sec:exp_rf}

\begin{figure}[t]
    \centering
    \begin{subfigure}[b]{\threecolfigwidth}
        \includegraphics[width=\textwidth]{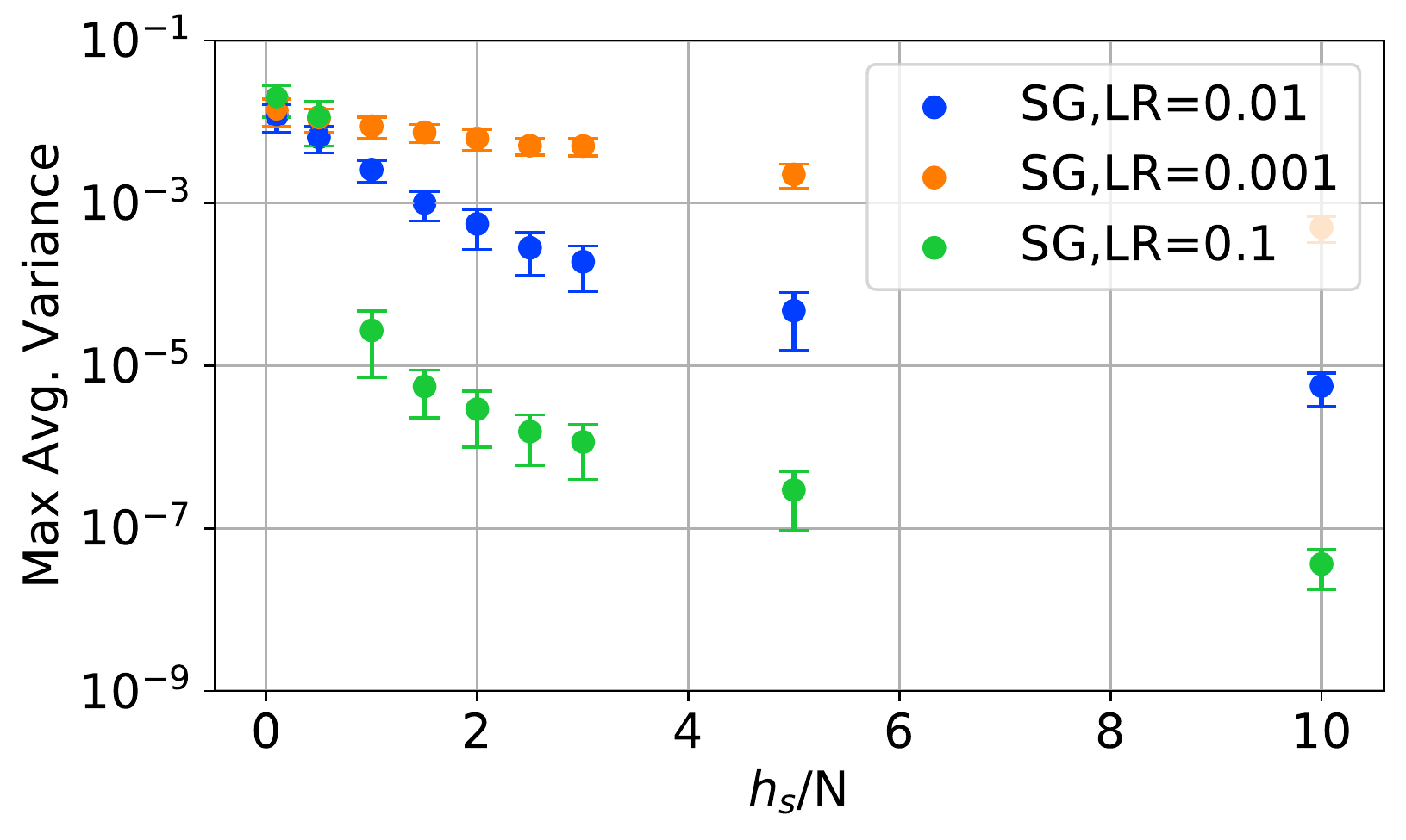}
        \vspace*{\capshift}
        \caption{$3$ SGD trajectories}
        \label{fig:rf_sgd_var}
    \end{subfigure}
    \begin{subfigure}[b]{\threecolfigwidth}
        \includegraphics[width=\textwidth]{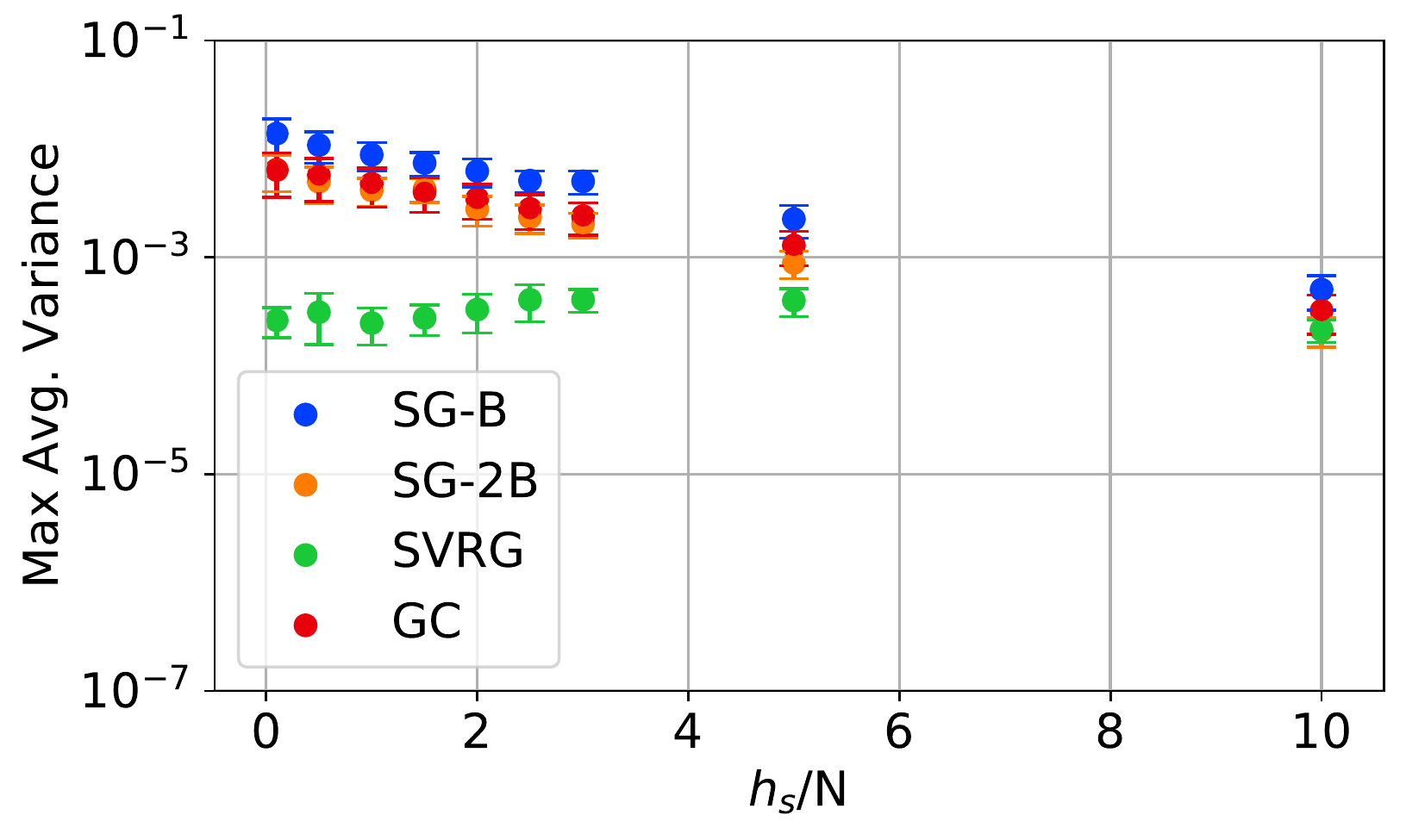}
        \vspace*{\capshift}
        \caption{Trajectory of LR=$0.001$}
        \label{fig:rf_lr0.001_var}
    \end{subfigure}
    \begin{subfigure}[b]{\threecolfigwidth}
        \includegraphics[width=\textwidth]{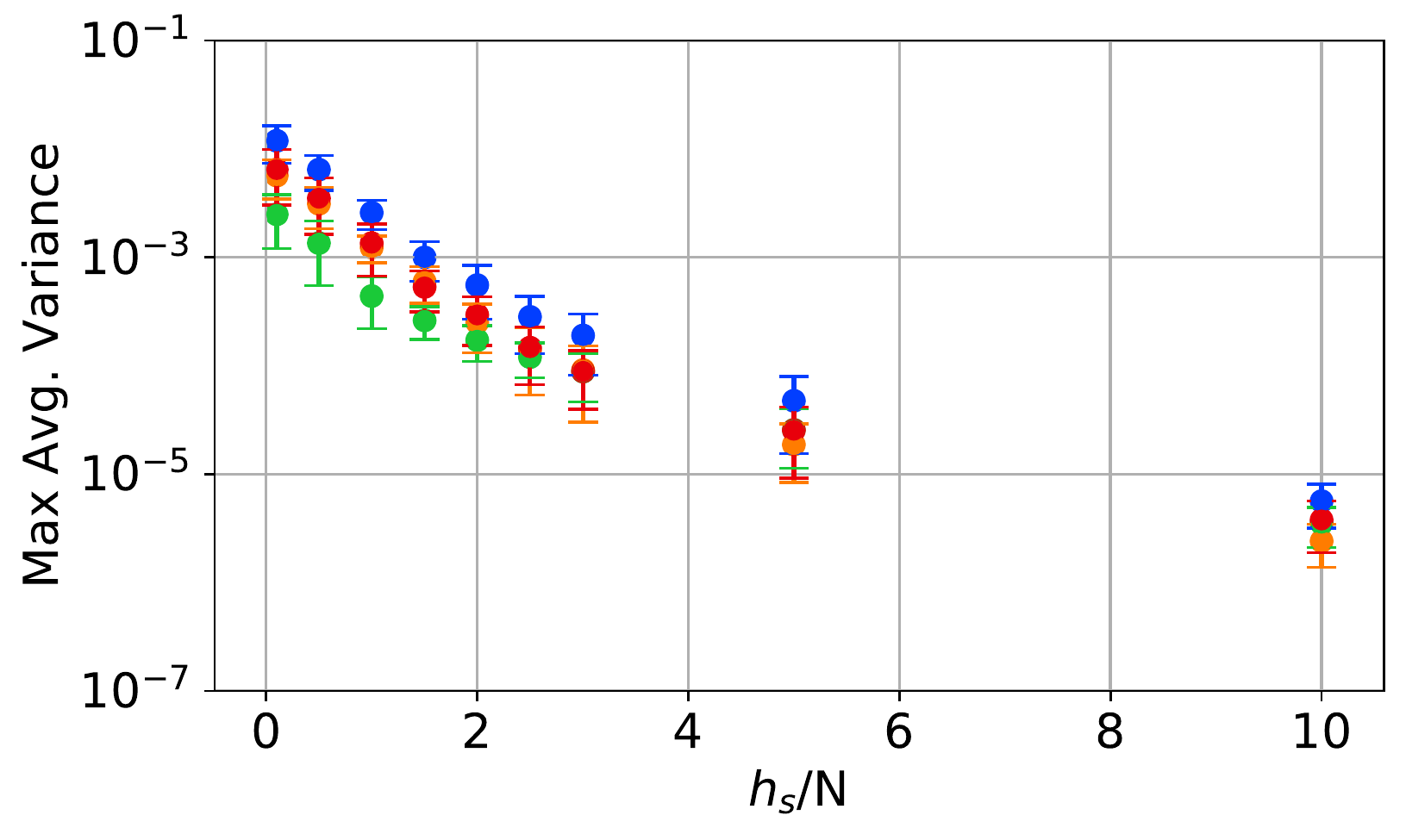}
        \vspace*{\capshift}
        \caption{Trajectory of LR=$0.01$}
        \label{fig:rf_lr0.01_var}
    \end{subfigure}
    \caption{{\bf Random Features models.}\/
    Variance (log-scale) versus the over-parametrization coefficient (student's 
    hidden divided by the training set size).
    We observe: teacher's hidden is not influential,
    variance is low in overparametrized regime,
    and with larger learning rates.
    We aggregate results from hyperparameters not shown.
    }
    \label{fig:rf}
    \vspace*{-12pt}
\end{figure}

The Random Features (RF) model~\citep{rahimi2007random} provides an effective 
way to explore the behaviour of optimization methods across a family of 
learning problems.
The RF model facilitates the discovery of optimization behaviours including 
the double-descent shape of the risk curve~\citep{hastie2019surprises, 
mei2019generalization}.
We train a student RF model with hidden dimensions $h_s$ on a fixed training 
set,
$(\bm{x}_i, \bm{y}_i)\in\R^I\times\{\pm1\}$,
$i=1,\ldots,N$,
sampled from a model,
${\bm{x}_i\sim \N(0, \I)}$,
${\bm{y}_i 
= \sign(\sigma(\bm{x}_i^\top\bm{\hat{\theta_1}})^\top\bm{\hat{\theta_2}}+b)}$
where $\sigma$ is the ReLU activation function, and the teacher hidden features 
$\bm{\hat{\theta_1}}\in\R^{I\times h_t}$, and second layer weights and bias, 
$\bm{\hat{\theta_2}}\in\R^{h_t\times 1}, b\in\R$, are sampled from the standard 
normal distribution.  Each $I$ dimensional random feature of the teacher is 
scaled to $\ell_2$ norm 1.  We train a student RF model with random features 
$\bm{\theta_1}\in\R^{I\times h_s}$ and second layer weights 
$\bm{\theta_2}\in\R^{h_s\times 1}$ by minimizing the cross-entropy loss.
In \cref{fig:rf}, we train hundreds of Random Features models and plot the 
average variance and normalized variance of gradient estimators.  We show both 
maximum and mean of the statistics during training.  The maximum better 
captures fluctuations of a gradient estimator and allows us to link our 
observations of variance to generalization using standard convergence bounds 
that rely on bounded noise~\citep{bottou2018optimization}.

\begin{figure}[t]
    \centering
    \begin{minipage}{\threecolfigwidth}
    \centering
   \begin{subfigure}[b]{\textwidth}
       \includegraphics[width=\textwidth]{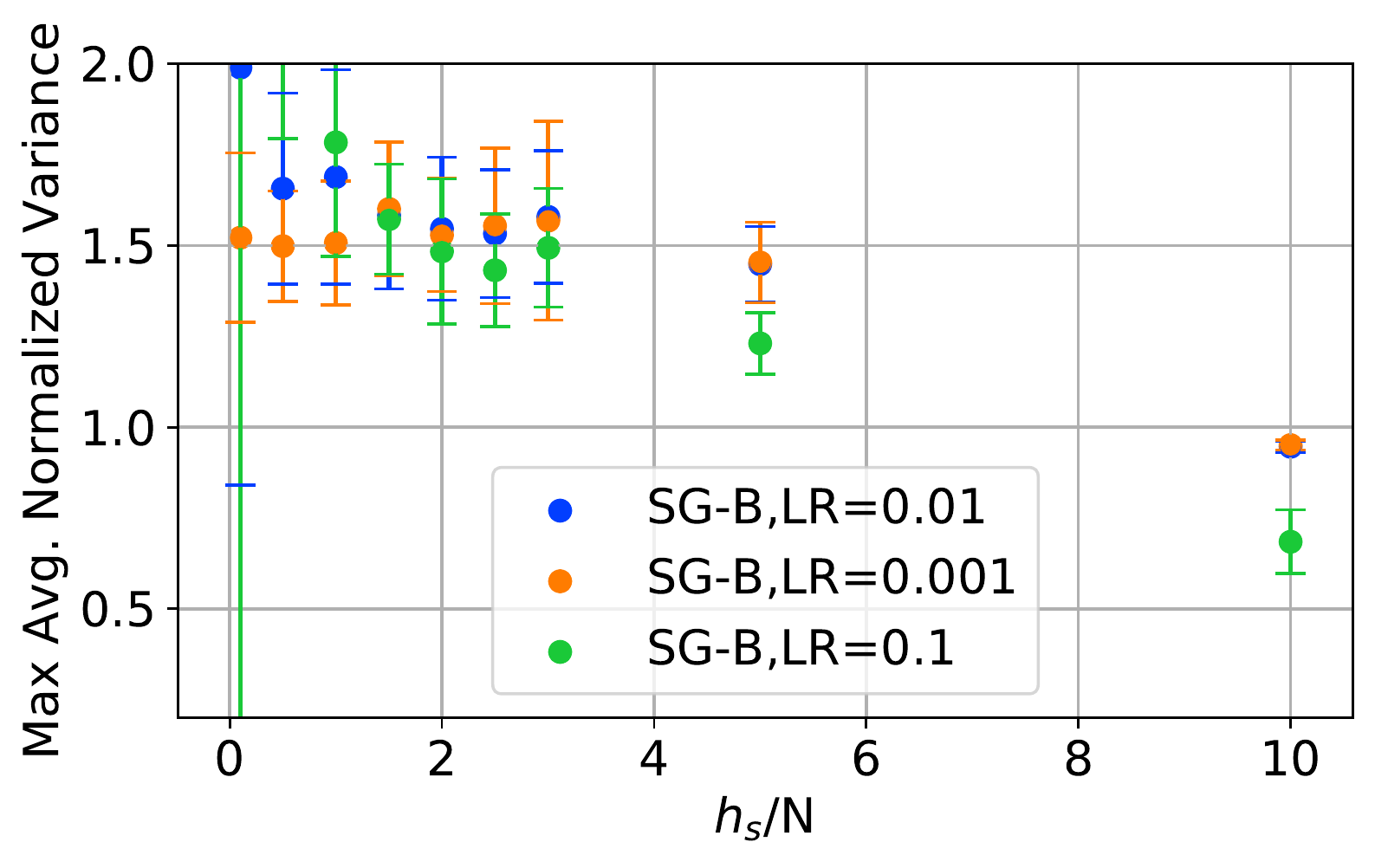}
       \vspace*{\capshift}
       \caption{SGD max norm.\ var.}
       \label{fig:rf_sgd_nvar_max}
   \end{subfigure}\\
   \begin{subfigure}[b]{\textwidth}
       \includegraphics[width=\textwidth]{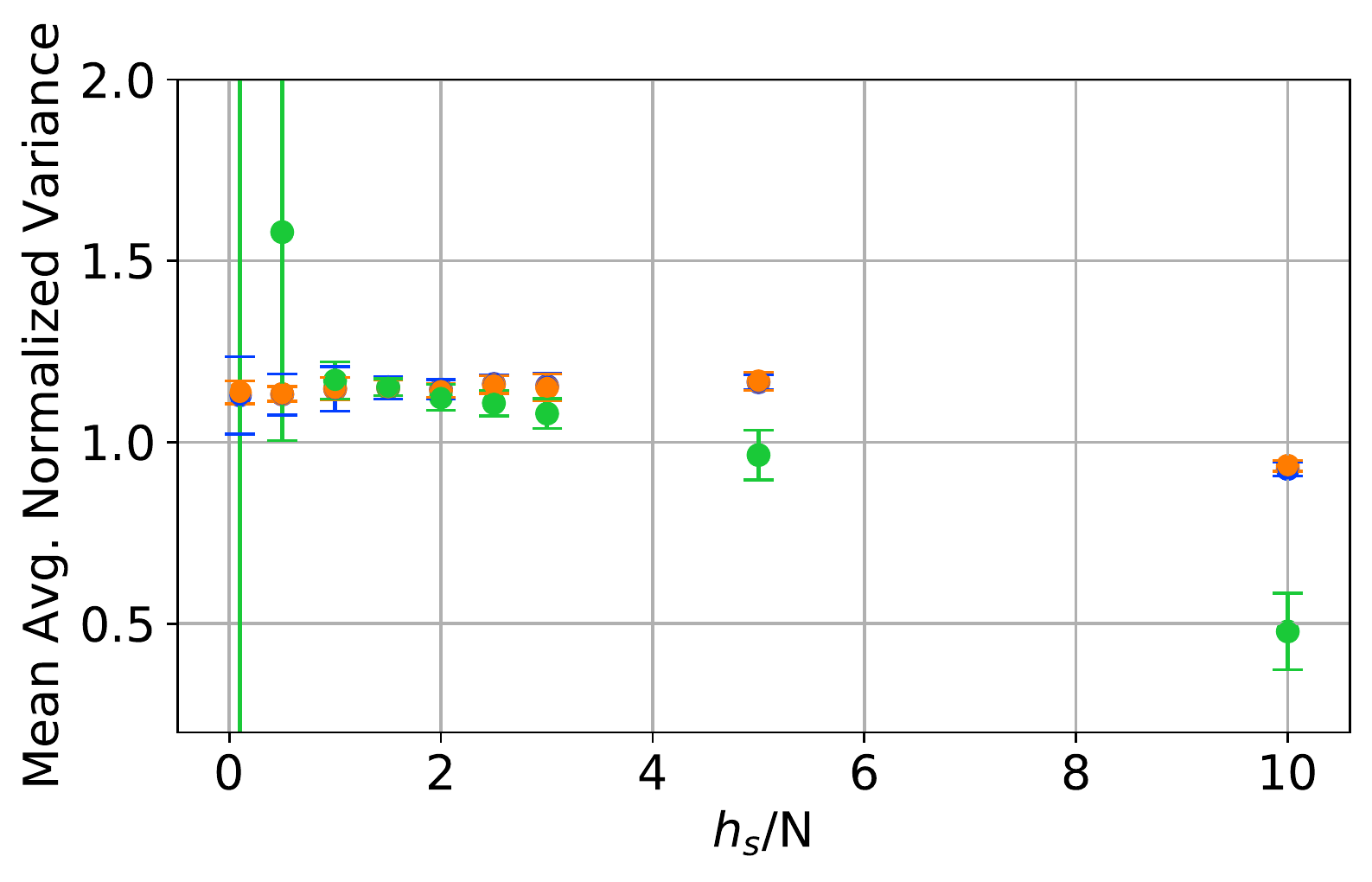}
       \vspace*{\capshift}
       \caption{SGD mean norm.\ var.}
       \label{fig:rf_sgd_nvar_mean}
   \end{subfigure}
    \caption{{\bf Normalized variance on overparam.\ RF} is less than $1$.}
    \label{fig:rf_sgd_nvar}
    \end{minipage}
    \hfill
    \begin{minipage}{0.32\textwidth}
    \centering
   \begin{subfigure}[b]{\textwidth}
       \includegraphics[width=\textwidth]{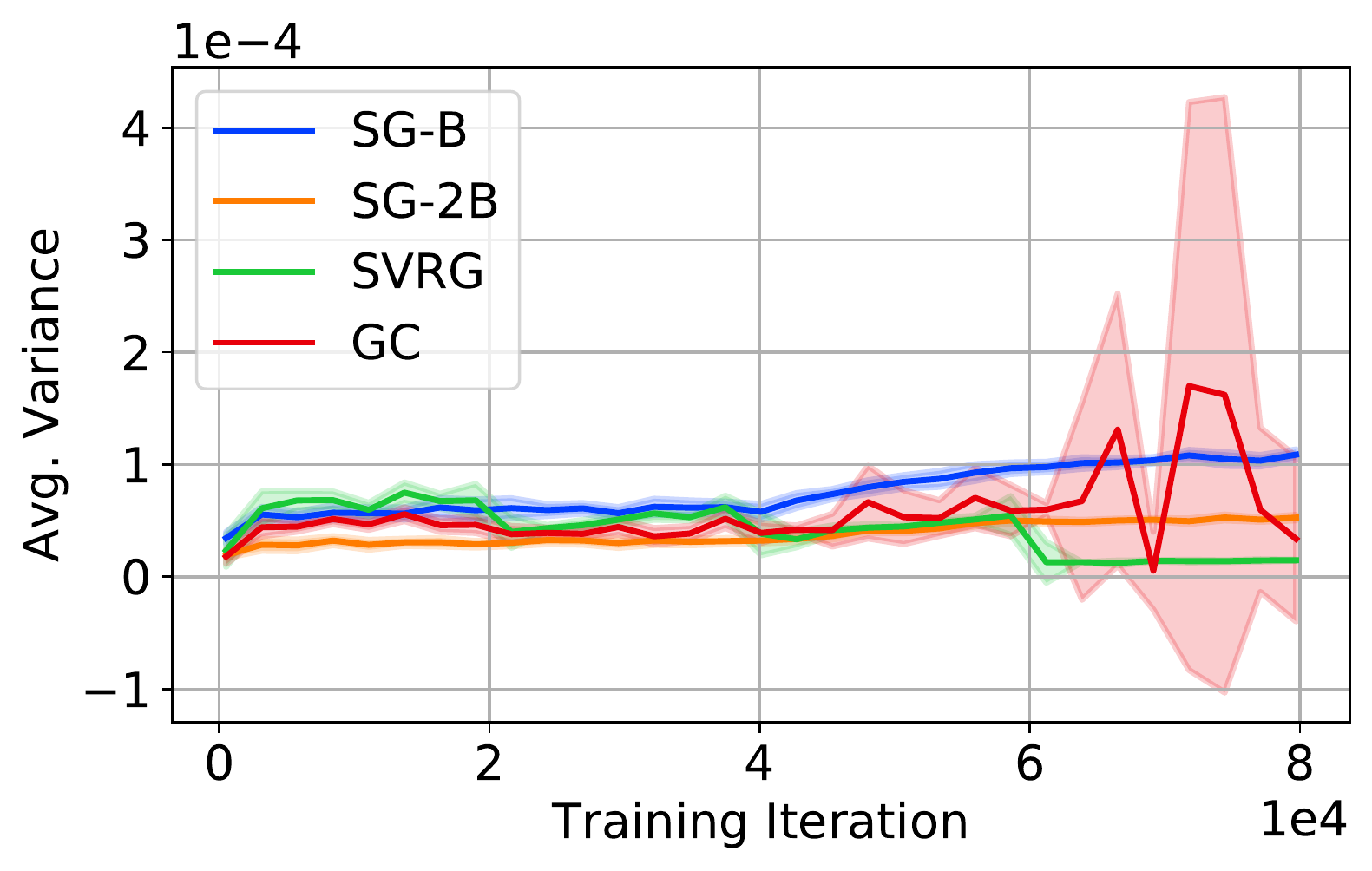}
       \vspace*{\capshift}
       \caption{Label smoothing}
       \label{fig:cifar10_label_smooth}
   \end{subfigure}\\
   \begin{subfigure}[b]{\textwidth}
       \includegraphics[width=\textwidth]{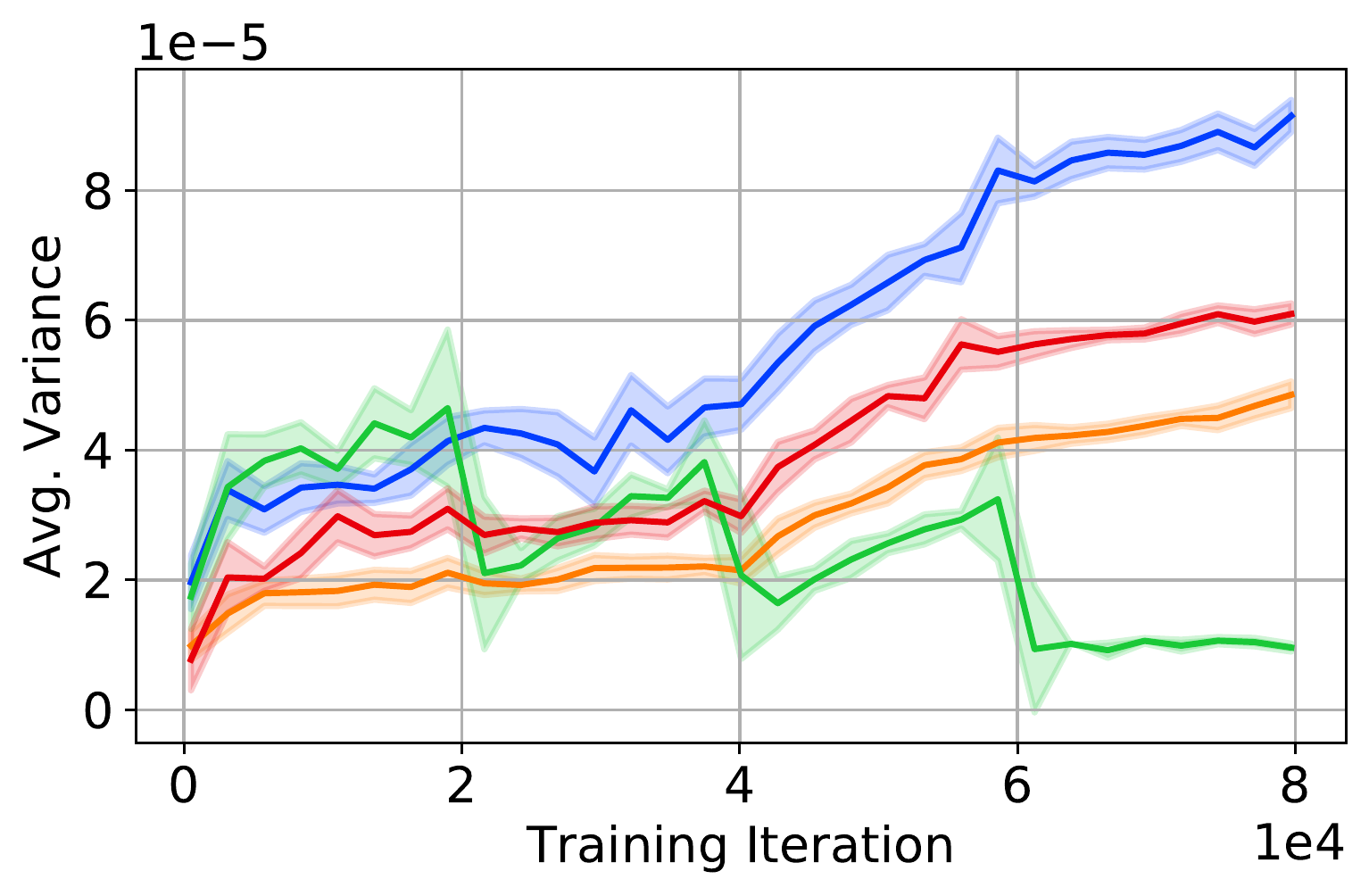}
       \vspace*{\capshift}
       \caption{Corrupt labels}
       \label{fig:cifar10_corrupt}
   \end{subfigure}
        \caption{{\bf CIFAR-10 Fluctuations} disappear with corrupt labels.}
    \label{fig:cifar10_noise}
    \end{minipage}
    \hfill
    \begin{minipage}{\threecolfigwidth}
        \centering
   \begin{subfigure}[b]{\textwidth}
        \includegraphics[width=\textwidth]{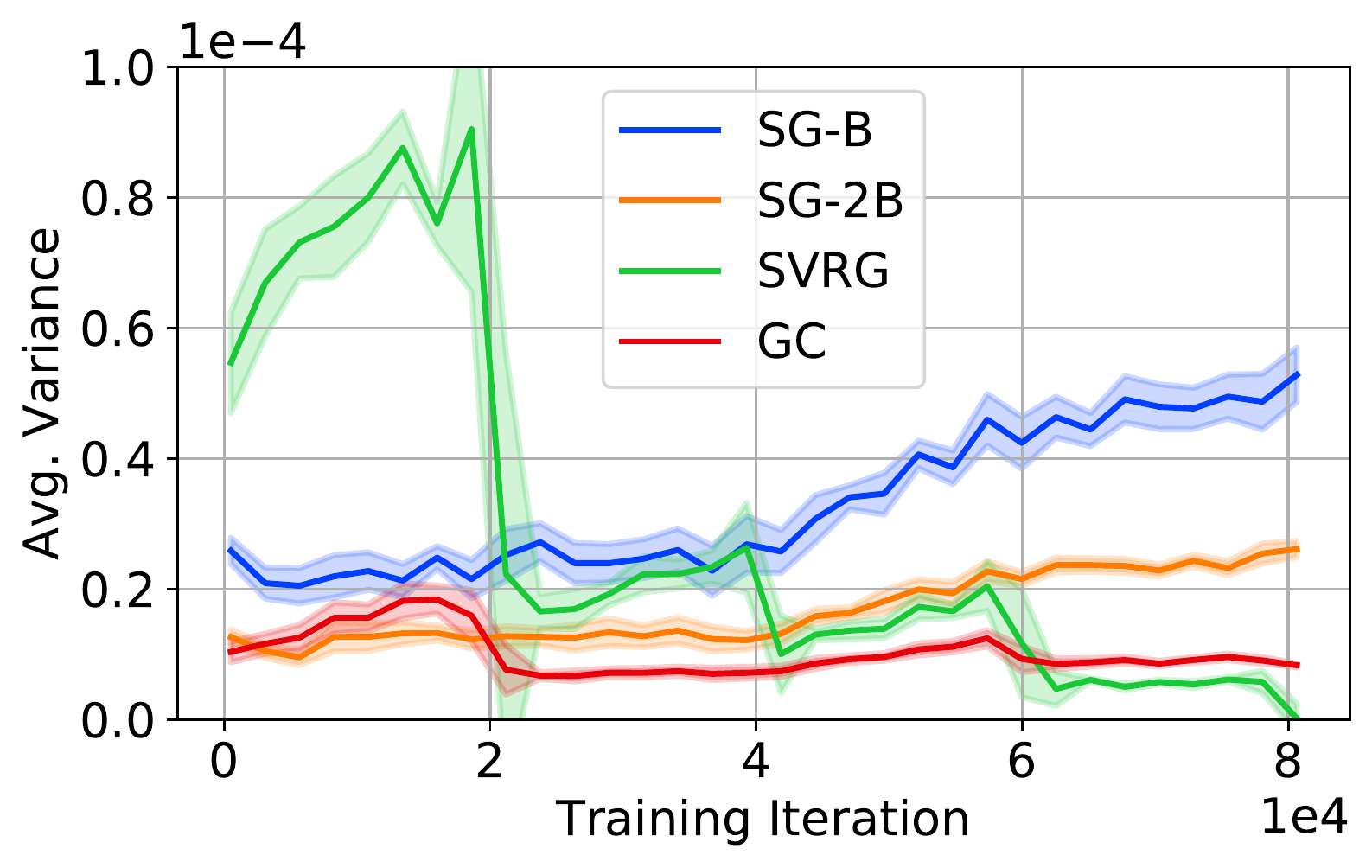}\\
       \vspace*{\capshift}
       \caption{CIFAR-10}
       \label{fig:cifar10_dup}
   \end{subfigure}
   \begin{subfigure}[b]{\textwidth}
        \includegraphics[width=\textwidth]{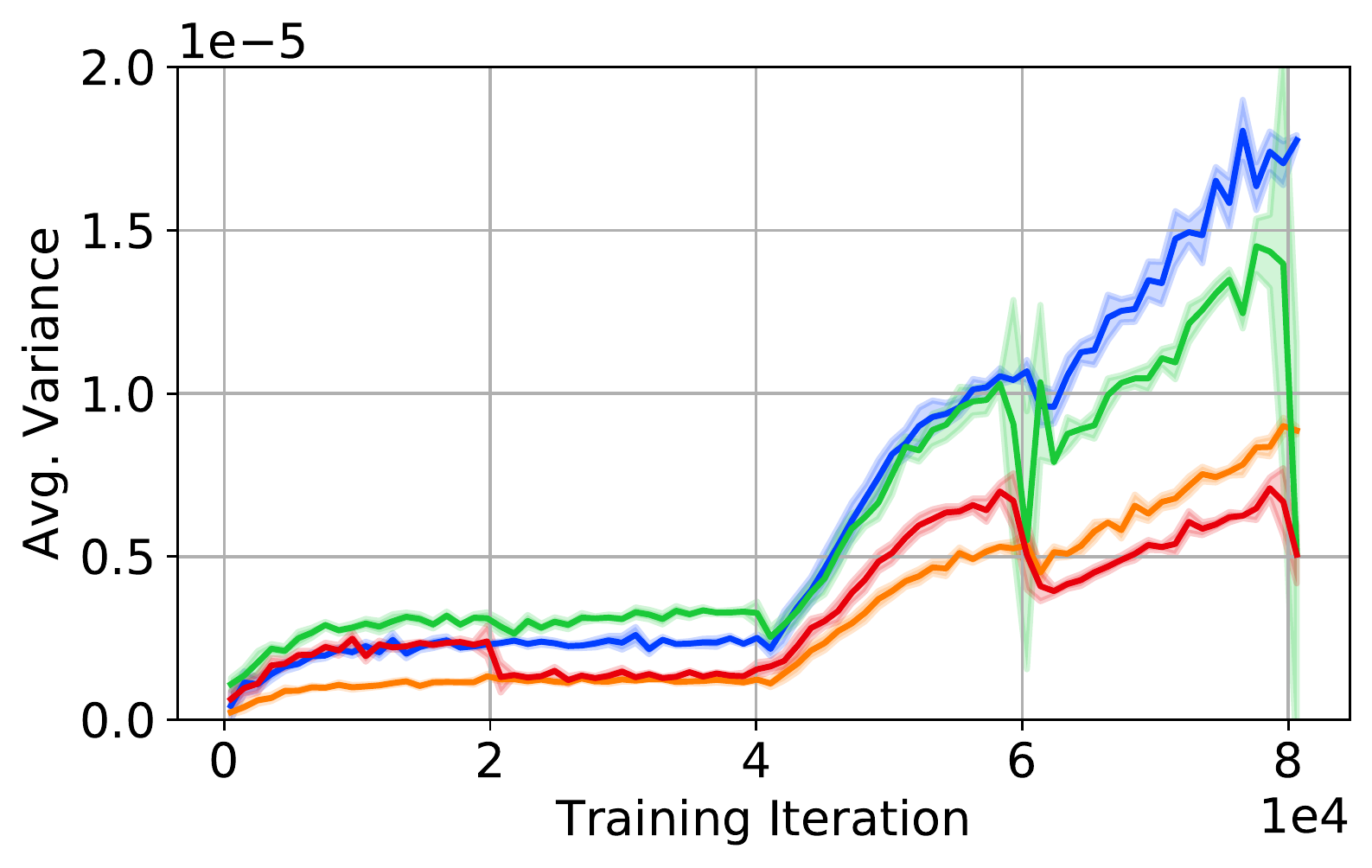}
       \vspace*{\capshift}
       \caption{CIFAR-100}
       \label{fig:cifar100_dup}
   \end{subfigure}
        \caption{{\bf Image classification with duplicates} exploited by \GC.}
        \label{fig:image_dup}
    \end{minipage}
    \vspace*{-14pt}
\end{figure}

{\bf Do models with small generalization gap converge faster?}\/
Based on small error bars, the only hyperparameters that affect the variance 
are learning rate and the ratio of the size of the student hidden layer over 
the training set size.  In contrast, in analysis of risk and the double descent 
phenomena, we usually observe a dependence on ratio of the student hidden layer 
size to the teacher hidden layer size~\citep{mei2019generalization}. This 
suggests that models that generalize better are not necessarily ones that train 
faster.

{\bf Does ``diminishing returns'' happen because of  overparametrization?}\/
\cref{fig:rf_lr0.001_var,fig:rf_lr0.01_var} show that with the same learning 
rate, all methods achieve similar variance in the overparametrized regime.  
Note that due to the normalization of random features, the gradients in each 
coordinate are expected to decrease as overparametrization increases. We 
conjecture that the diminishing returns in increasing the mini-batch size 
should also be observed in overparametrized random features models similar to 
linear and deeper models~\citep{zhang2019algorithmic, shallue2018measuring}.

{\bf Why does the loss usually drop immediately after a learning rate drop?}
\cref{fig:rf_sgd_var} shows that the variance is smaller for trajectories with 
larger learning rates and that the gap grows as overparametrization grows.  
This is a direct consequence of the dependence of the noise in the gradient on 
current parameters. In \cref{sec:exp_image} we observe the opposite of this 
behaviour in deep models.   In contrast, \cref{fig:rf_sgd_nvar} shows that for 
overparametrization less than $5$, all trajectories have similar normalized 
variance that is larger than one (noise is more powerful than the gradient).  
As such, we hypothesize that the reduction in variance is not the sole reason 
for the immediate decrease in the loss after a learning rate 
drop~\citep[e.g.][]{he2016deep}.

{\bf How does SGD avoid local minima?}
In \cref{fig:rf_sgd_nvar_max}, the error bars in the maximum normalized 
variance are long for overparametrization coefficient less than $5$.  The 
reason is in some iterations the second moment of the gradient for some 
coordinates gets close to zero but the noise due to mini-batching is still 
non-zero.  Often in the next iteration, the gradient becomes large again.  This 
is an example where SGD avoids local minima due to noise.

{\bf Why does SVRG fail in deep learning?}
\cref{fig:rf_lr0.001_var,fig:rf_lr0.01_var} show that the gain of SVRG vanishes 
in the over-parametrized regime ($10$) where all methods have relatively low 
variance (below $10^{-5}$). We hypothesize that the cause is the generally 
lower variance of the noise in the overparametrized regime rather than the 
staleness of the control-variate.

\subsection{Duplicates: Back to the Motivation for Gradient Clustering}
\begin{figure}[t]
    \centering
    \begin{subfigure}[b]{\threecolfigwidth}
        \includegraphics[width=\textwidth]{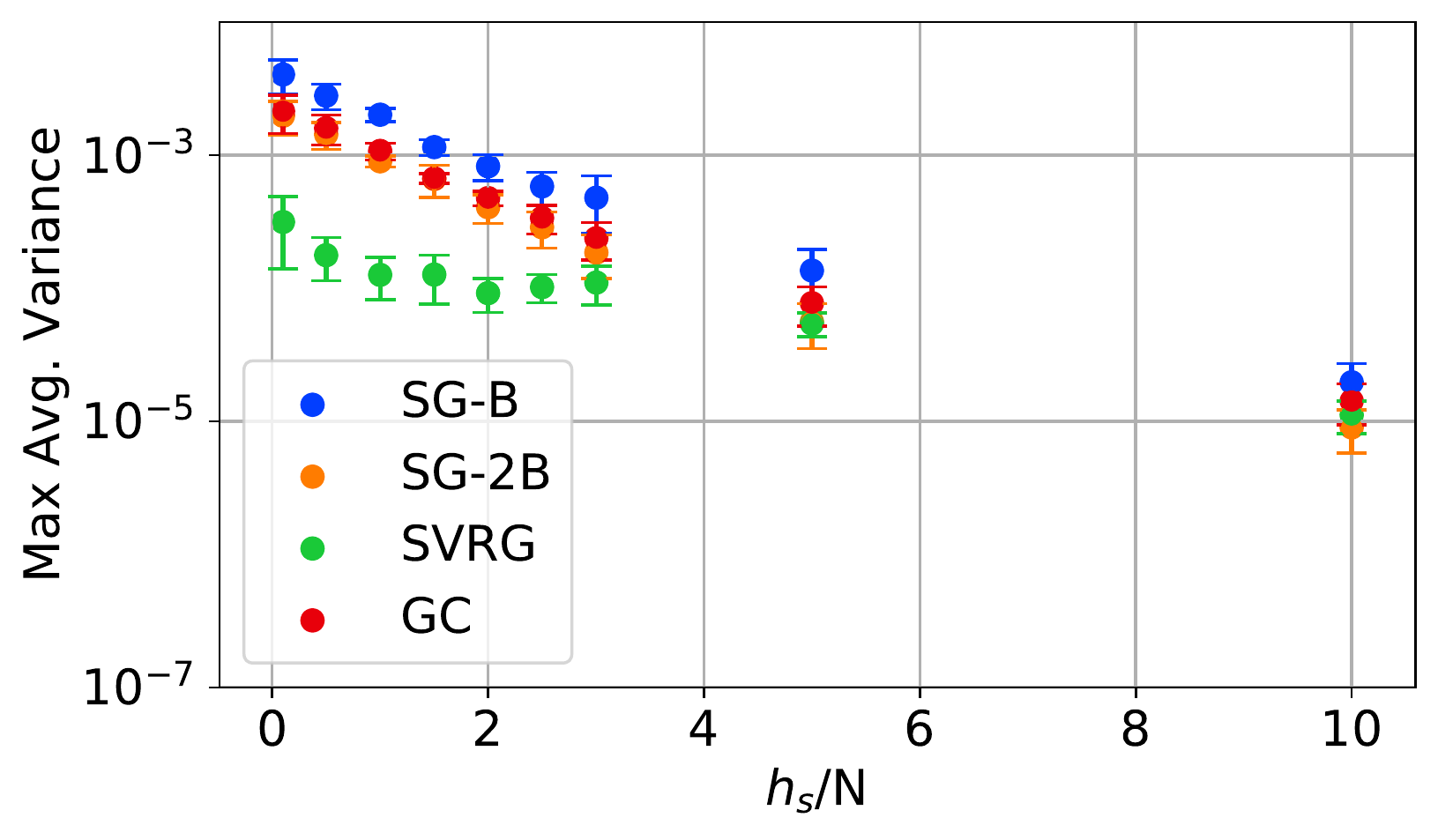}
        \vspace*{\capshift}
        \caption{$10\%$ duplicates}
        \label{fig:dup_10}
    \end{subfigure}
    \begin{subfigure}[b]{\threecolfigwidth}
        \includegraphics[width=\textwidth]{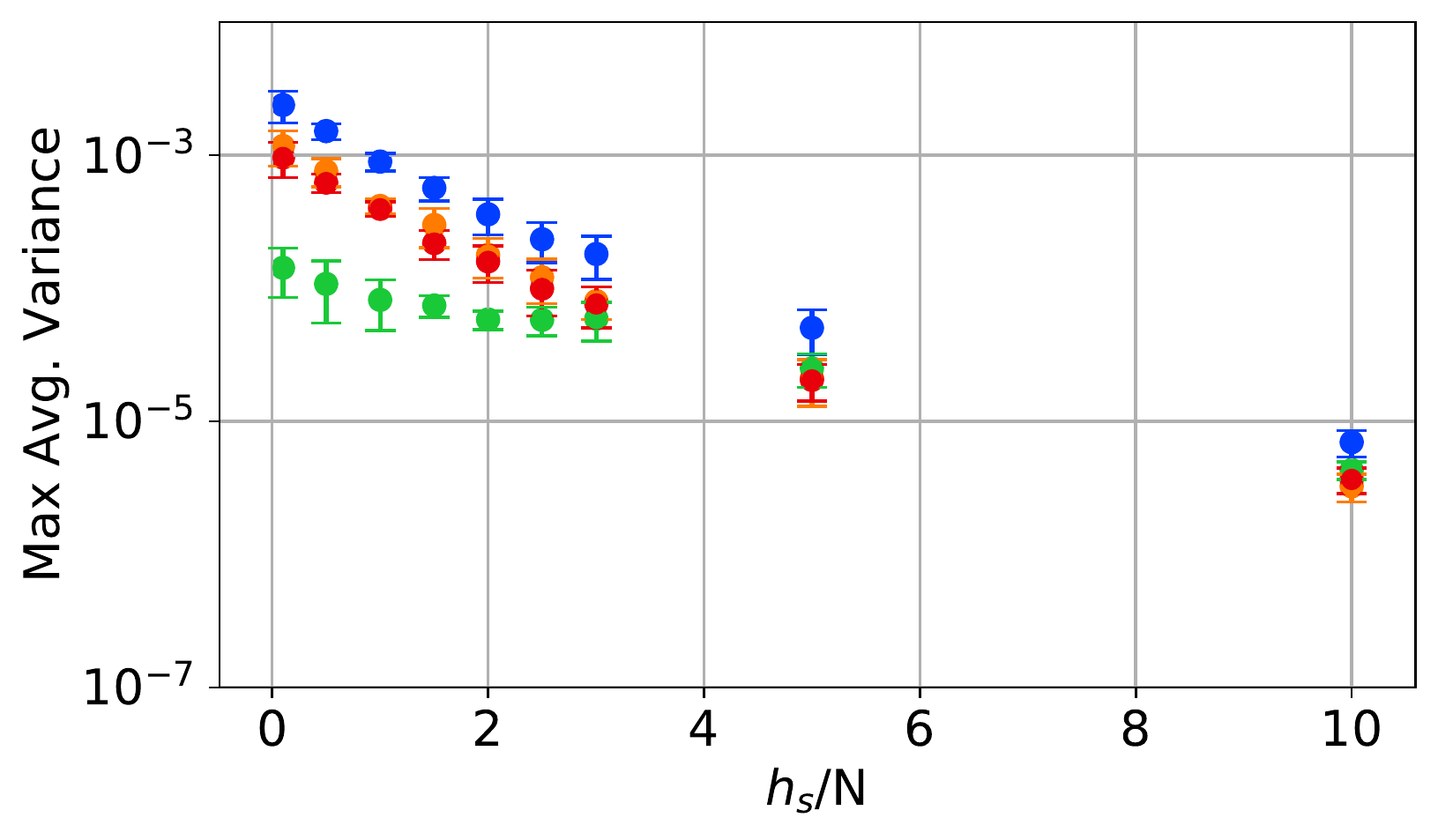}
        \vspace*{\capshift}
        \caption{$50\%$ duplicates}
        \label{fig:dup_50}
    \end{subfigure}
    \begin{subfigure}[b]{\threecolfigwidth}
        \includegraphics[width=\textwidth]{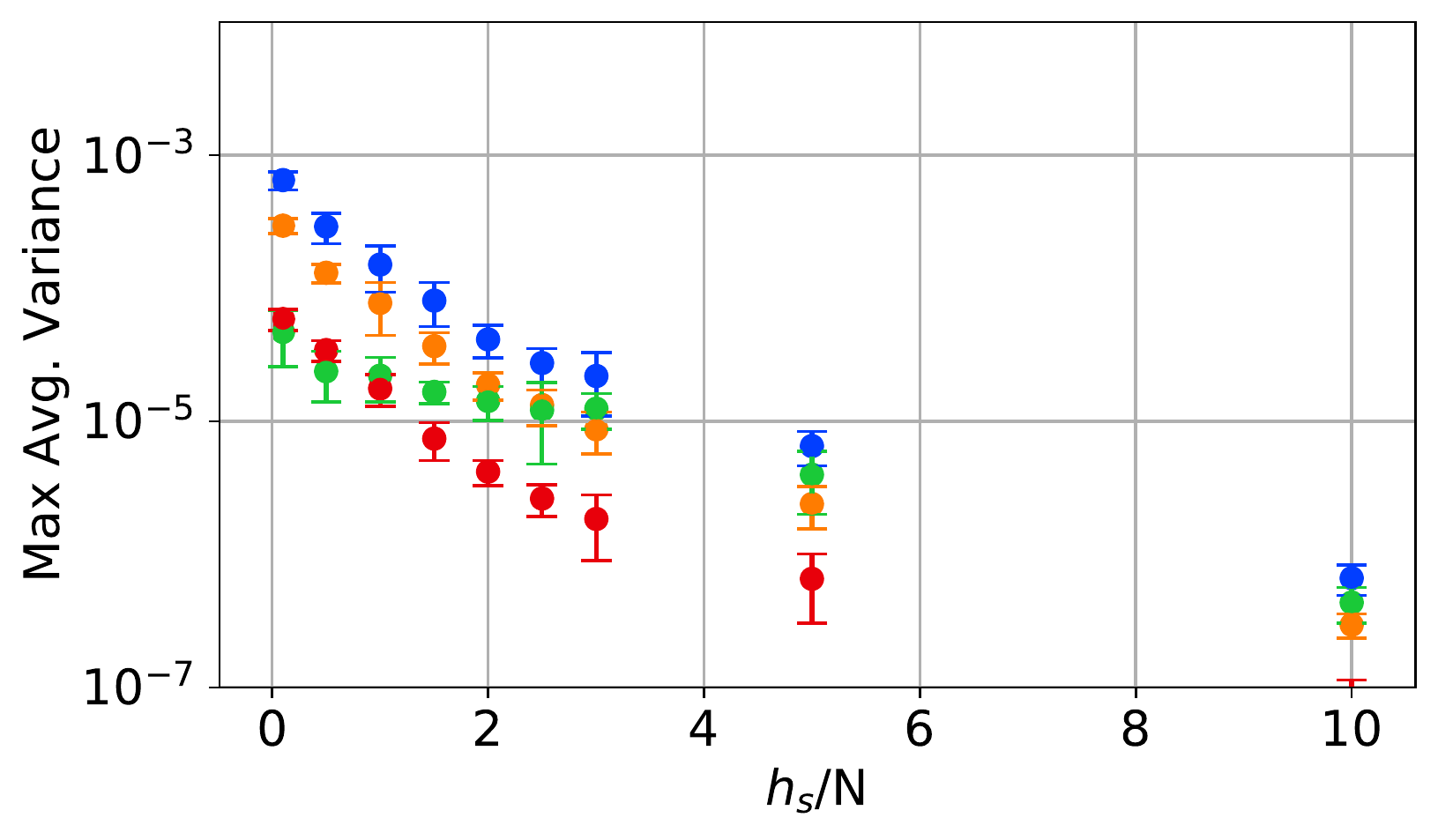}
        \vspace*{\capshift}
        \caption{$90\%$ duplicates}
        \label{fig:dup_90}
    \end{subfigure}
    \caption{{\bf Training RF Models with Duplicates.}\/ \GC identifies and 
    exploits duplicates.  Plots are similar to \cref{fig:rf}.  Learning rate in 
    all three is $0.01$.  In each training, there are $5$ data points that are 
    repeated equally to make up $10\%$ (left), $50\%$ (middle), and $90\%$ 
    (right) of the training set.
    }
    \label{fig:rf_dup}
    \vspace*{-14pt}
\end{figure}

In \cref{fig:rf_dup}, we trained random features models with additional 
duplicated data points.  We observe that as the ratio of duplicates to 
non-duplicates increases, the gap between the variance of \Gluster and other 
methods improve.  Without duplicate data, \Gluster is always between \SG and 
\SGdB.  It is almost never worse than \SG and never better than \SGdB.  
\Gluster is as good as \SGdB at mild overparametrization ($1-4$). We need 
a degree of overparametrization for \Gluster to reduce the variance but too 
much overparametrization leaves no room for improvement. When duplicates exist, 
\Gluster performs well with a gap that does not decrease by 
overparametrization.

Similarly, experiments on CIFAR-10 and CIFAR-100 (\cref{fig:image_dup}) show 
that \Gluster significantly reduces the variance when duplicate data points 
exist.  Note that because of common data augmentations, duplicate data points 
are not exactly duplicate in the input space and there is no guarantee that 
their gradients would be similar.

\section{Conclusion}
In this work we introduced tools for understanding the optimization behaviour 
and explaining previously perplexing observations in deep learning. We expect 
our contributions to not only guide improvements to optimization speed and 
generalization but also the design of interpretable models.
We have provided evidence that structured gradient distributions such as 
clustered gradients exist and that statistics of gradients can provide insight 
into the optimization performance.  However, exploiting this knowledge to 
improve optimization has proven to be challenging.

\begin{ack}
    The authors would like to thank Nicolas Le Roux, Fabian Pedregosa, Mark 
    Schmidt, Roger Grosse, Sara Sabour, and Aryan Arbabi for helpful 
    discussions and feedbacks on this manuscript.
    Resources used in preparing this research were provided, in part, by the 
    Province of Ontario, the Government of Canada through CIFAR, and companies 
    sponsoring the Vector Institute.
\end{ack}

\bibliography{bib}
\bibliographystyle{plainnat}

\vfill

\pagebreak
\appendix
\section{Additional Details of Gradient Clustering (\cref{sec:gvar})}
    \subsection{Proof of \cref{thm:gvar}}\label{sec:gvar_proof}
    The gradient estimator, $\gC$, is unbiased for any partitioning of data, i.e.\ 
equal to the average gradient of the training set,
\begin{align}
    \E[\gC]
    &
    = \frac{1}{N} \SK \Nk \E[\gjk]
    = \frac{1}{N} \SK \Nk \left(\frac{1}{\Nk} \SJ \gjk\right)
    = \underbrace{\frac{1}{N} \SK \SJ \gjk}_{(*)}
    = \frac{1}{N} \SI \gi = \bm{g} ~,\nonumber
\end{align}
where we use the fact that the expectation of a random sample drawn uniformly 
from a subset is equal to the expectation of the average of samples from that 
subset.  Also note that the gradient of every training example appears once in 
$(*)$. 

Although partitioning does not affect the bias of $\gC$, it does affect the 
variance,
\begin{align}
    \V[\gC]
    &= \frac{1}{N^2} \left(\SI \Nk^2 \V[\gjk]
    + 2\SK\sum_{\hat{k}=1}^{N_{\hat{k}}}
    \Nk N_{\hat{k}} \C[\gjk, \bm{g}_{\hat{j},\hat{k}}] \right) 
    = \frac{1}{N^2} \SK \Nk^2 \V[\gjk]
\end{align}
where the variance is defined as the trace of the covariance matrix. Since we 
assume the training set is sampled i.i.d., the covariance between gradients of 
any two samples is zero.  In a dataset with duplicate samples, the gradients of 
duplicates will be clustered into one cluster with zero variance if mingled 
with no other data points.

\section{Additional Details of Efficient \GC (\cref{sec:gluster})}
    \subsection{Convolutional Layers}\label{sec:conv}
        
In neural networks, the convolution operation is performed as an inner product 
between a set of weights
$\lparam\in \R^{h\times w\times \hat{\Di}\times \Do}$,
namely kernels, by patches of size
$h\times w$
in the input.  Assuming that we have preprocessed the input by extracting 
patches, the gradient w.r.t.\ $\lparam$ is
    $ \gb = {\sum_t \bm{g}_{b,t} }$,
$\bm{g}_{b,t}\in \R^{\Di\times \Do}$ is the gradient at the spatial location 
$t\in T$ and
$\Di=h \times w \times \hat{\Di}$
is the flattened dimension of a patch.
The gradient at spatial location $t$ is computed as $\bm{g}_{b,t}= \AAA_{b,t} 
\DDD_{b,t}^\top$.

Like the fully-connected case, we use a rank-$1$ approximation to the cluster 
centers in a convolution layer, defining $\Ck=\ck \dk^\top$.  As such, \bAU steps 
are performed efficiently. For the \bA step we rewrite
${\vv{\Ck}\odot \vv{\gb}}$,
\begin{align}
    \vv{\Ck}\odot \vv{\sum_t \AAA_{b,t} \DDD_{b,t}^\top}
    &= \sum_{u,v} (\bm{c}_{ku} ~ \bm{d}_{kv} ~~ (\sum_t \AAA_{btu} ~ \DDD_{btv}))
    \label{eq:conv1}\\
    &= \sum_t (\sum_u \bm{c}_{ku} ~ \AAA_{btu}) (\sum_{v} \bm{d}_{kv} 
    ~ \DDD_{btv})
    ,
    \label{eq:conv2}
\end{align}
where the input dimension is indexed by $u$ and the output dimension is indexed 
by $v$. Eqs.~\ref{eq:conv1} and~\ref{eq:conv2} provide two ways of computing 
the inner-product, where we first compute the inner sums, then the outer sum. The 
efficiency of each formulation depends on the size of the kernel and layer's 
input and output dimensions.

    \subsection{Complexity Analysis}\label{sec:comp}
        {\renewcommand{\arraystretch}{1.5}
\begin{table}[t]
    \centering
    \begin{tabular}{ccc}
        \toprule
        {\bf Operation} & {\bf FC Complexity} & {\bf Conv Complexity}\\
        \midrule
        $\bm{C}\odot \bm{g}$ & $K B (\Di+\Do)$
        & \makecell{%
            Eq.~\ref{eq:conv1}: $B(T+K)\Di \Do$\\
         Eq.~\ref{eq:conv2}: $ B T K (\Di + \Do)$}\\
        $\bm{C}\odot \bm{C}$ & $K (\Di+\Do)$ & $K (\Di+\Do)$\\ 
        $\bm{g}\odot \bm{g}$ & $B (\Di+\Do)$
        & \makecell{%
            Eq.~\ref{eq:conv1}: $B T \Di \Do,$ \\
        Eq.~\ref{eq:conv2}: $B T^2 (\Di + \Do)$}\\
        \midrule
        Back-prop & $B \Di \Do$ & $B T \Di \Do$\\
        \bA step & $K B (\Di+\Do)$ & See Sec.~\ref{sec:comp}\\
        \bU step & $B (\Di + \Do)$ & $B(\Di +\Do)$\\
        \bottomrule
    \end{tabular}
    \caption{Complexity of \Gluster compared to the cost of back-prop.}
    \label{tab:comp}
\end{table}
}

\Gluster, described in \cref{alg:full}, performs two sets of operations, 
namely, the cluster center updates (\bU step), and the assignment update of 
data to clusters (\bA step). \bA steps instantly affect the optimization by 
changing the sampling process.  As such, we perform an \bA step every few 
epochs and change the sampling right after. In contrast, the \bU step can be 
done in parallel and more frequently than the \bA step, or online using 
mini-batch updates. The cost of both steps is amortized over optimization 
steps.

Table~\ref{tab:comp} summarizes the run-time complexity of \Gluster compared to
the cost of single SGD step.
The \bU step is always cheaper than a single back-prop step.
The \bA step is cheaper for fully-connected layers if ${K<\min(\Di,\Do)}$.

For convolutional layers, we have two ways to compute the terms in the \bA step 
(Eqs.~\ref{eq:conv1} and~\ref{eq:conv2}).  For ${\bm{C}\odot \bm{g}}$, if
${\min(T, K)}<{K<\min(\Di, \Do)}$, Eq.~\ref{eq:conv2} is more efficient. For 
${\bm{g}\odot \bm{g}}$, Eq.~\ref{eq:conv2} is more efficient if $T<\min(I, O)$.  
If $K<T$, both methods have lower complexity than a single back-prop step. If 
we did not have the $\Nk$ multiplier in the \bA step, we could ignore the 
computation of the norm of the gradients, and hence further reduce the cost.

In common neural network architectures, the condition $K<T$ is easily satisfied 
as $T$ in all layers is almost always more than $10$ and usually greater than $100$, 
while $10$-$100$ clusters provides significant variance reduction. As such, the 
total overhead cost with an efficient implementation is at most $2\times$ the 
cost of a normal back-prop step. We can further reduce this cost by performing 
\Gluster on a subset of the layers, e.g.\ one might exclude the lowest 
convolutional layers.

The total memory overhead is equivalent to increasing the mini-batch size by 
$K$ samples as we only need to store rank-$1$ approximations to the cluster 
centers.

\section{Additional Details for Experiments (\cref{sec:exp})}\label{app:exp}

The mini-batch size in \Gluster and SVRG and the number of clusters in \Gluster 
are the same as the mini-batch size in \SG and the same as the mini-batch size 
used for training using SGD\@. To measure the gradient variance, we take 
snapshots of the model during training, sample tens of mini-batches from the 
training set (in case of \Gluster, with stratified sampling), and measure the 
average variance of the gradients.

We measure the performance metrics (e.g.\ loss, accuracy and variance) as  
functions of the number of training iterations rather than wall-clock time. In 
other words, we do not consider computational overhead of different methods.  
In practice, such analysis is valid as long as the additional operations could 
be parallelized with negligible cost.

\subsection{Experimental Details for Image Classification Models 
(\cref{sec:exp_image})}\label{app:exp_image}

On MNIST, our MLP model consists of there fully connected layers:
layer1: $28*28\times 1024$,
layer2: $1024\times 1024$,
layer3: $1024, 10$. We use ReLU activations and no dropout in this MLP\@. We 
train all methods with learning rate $0.02$, weight decay $5 \times 10^{-4}$, 
and momentum $0.5$.
On CIFAR-10, we train ResNet8 with no batch normalization layer and learning 
rate $0.01$, weight decay $5 \times 10^{-4}$, and momentum $0.9$ for $80000$ 
iterations.  We decay the learning rate at $40000$ and $60000$ iterations by 
a factor of $0.1$.
On CIFAR-100, we train ResNet32 starting with learning rate $0.1$. Other 
hyper-parameters are the same as in CIFAR-10.
On ImageNet, we train ResNet18 starting with learning rate $0.1$, weight decay 
$1\times 10^{-4}$, and momentum $0.9$. We use a similar learning rate schedule 
to CIFAR-10.

{\renewcommand{\arraystretch}{1.5}
\begin{table}[t]
    \centering
    \begin{tabular}{ccccccccc}
        \toprule
        Dataset & Model & $B$ & T & Log T
        & Estim T & U & \GC T\\
        \midrule
        MNIST & MLP & $128$ & $50000$ & $500$ & $50$ & $2000$ & $10$\\
        CIFAR-10 & ResNet8 & $128$ & $80000$ & $500$ & $50$ & $20000$ & $3$\\
        CIFAR-100 & ResNet32 & $128$ & $80000$ & $500$ & $50$ & $20000$ & $3$\\
        ImageNet & ResNet18 & $128$ & $80000$ & $1000$ & $10$ & $10000$ & $3$\\
        \bottomrule
    \end{tabular}
    \label{tbl:exp_setup}
    \caption{Hyperparameters.}
\end{table}
}

In \cref{tbl:exp_setup} we list the following hyperparameters: the interval of 
measuring gradient variance and normalized variance (Log T), number of gradient 
estimates used on measuring variance (Estim T), the interval of updating the 
control variate in SVRG and the clustering in \GC (U), and the number of \GC 
update iterations (\GC T).

In plots for random features models, each point is generated by keeping $h_s$ 
fixed at $1000$ and varying $N$ in the range $[0.1, 10]$. We average over $3$ 
random seeds, $2$ teacher hidden dimensions and $2$ input dimensions (both 
$\times0.1$ and $\times10$ student hidden). We use mini-batch size $10$ for 
\SG, SVRG, and \Gluster.

A rough estimate of the overparametrization coefficient (discussed in 
\cref{sec:exp_rf}) for deep models is to divide the total number of parameters 
by the number of training data. On MNIST the coefficient is approximately $37$ 
for CNN and $31$ for MLP\@. On CIFAR-10 it is approximately $3$ for ResNet8 and 
$9$ for ResNet32\@.  Common data augmentations increase the effective training 
set size by $10\times$. On the other hand, the depth potentially increases the 
capacity of models exponentially (cite the paper that theoretically says how 
many data points a model can memorize). As such, it is difficult to directly 
relate these numbers to the behaviours observed in RF models.

\subsection{Experimental Details for Random Features Models 
(\cref{sec:exp_rf})}\label{app:exp_rf}

The number of training iterations is chosen such that the training loss has 
flattened.  The maximum is taken over the last $70\%$ of iterations (the 
variance is usually high for all methods in the first $30\%$).  
Mean variance plots for random features models are similar to max variance 
plots presented in \cref{sec:exp_rf}.
\begin{figure}[t]
    \centering
    \includegraphics[width=.32\textwidth]{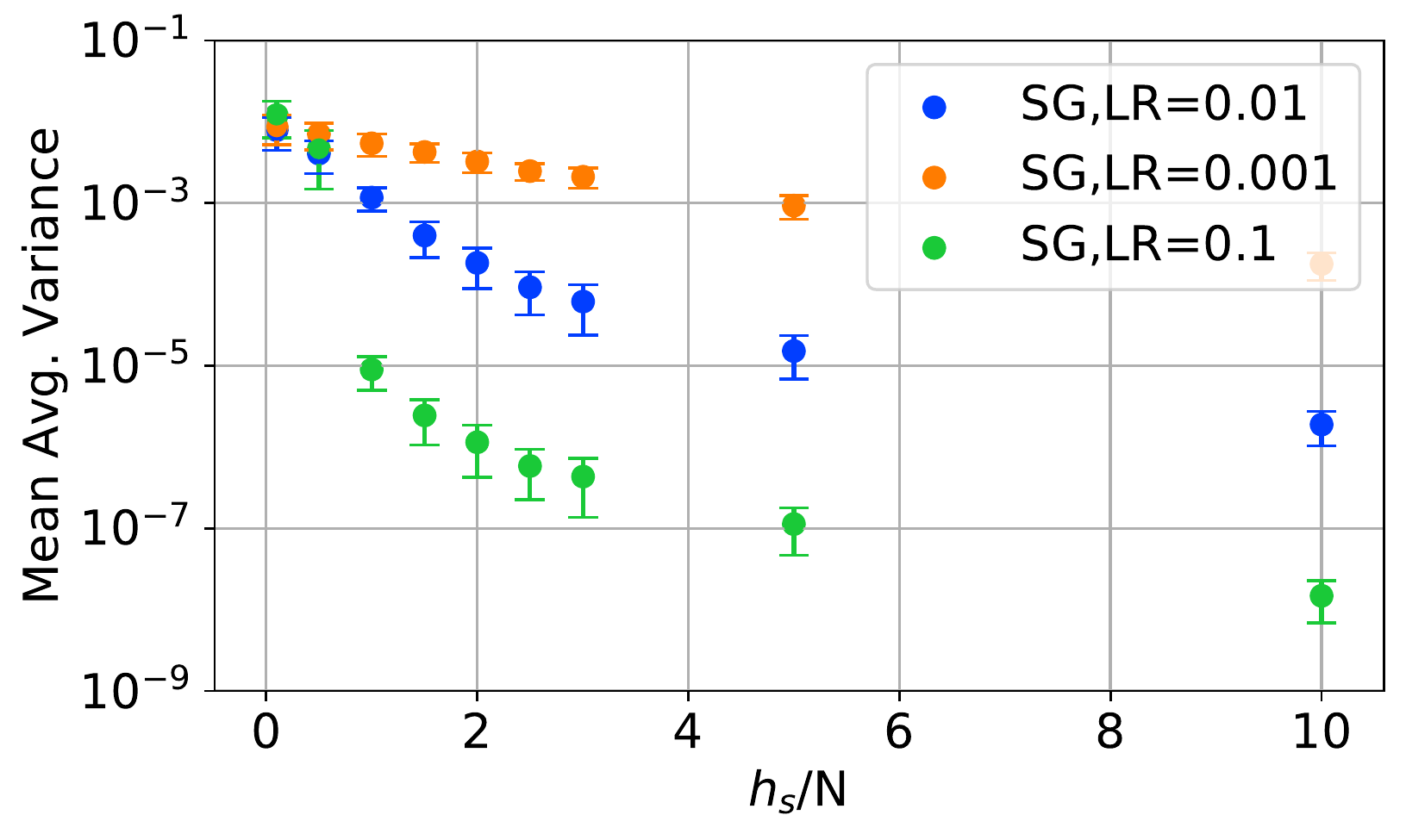}
    \hfill
    \includegraphics[width=.32\textwidth]{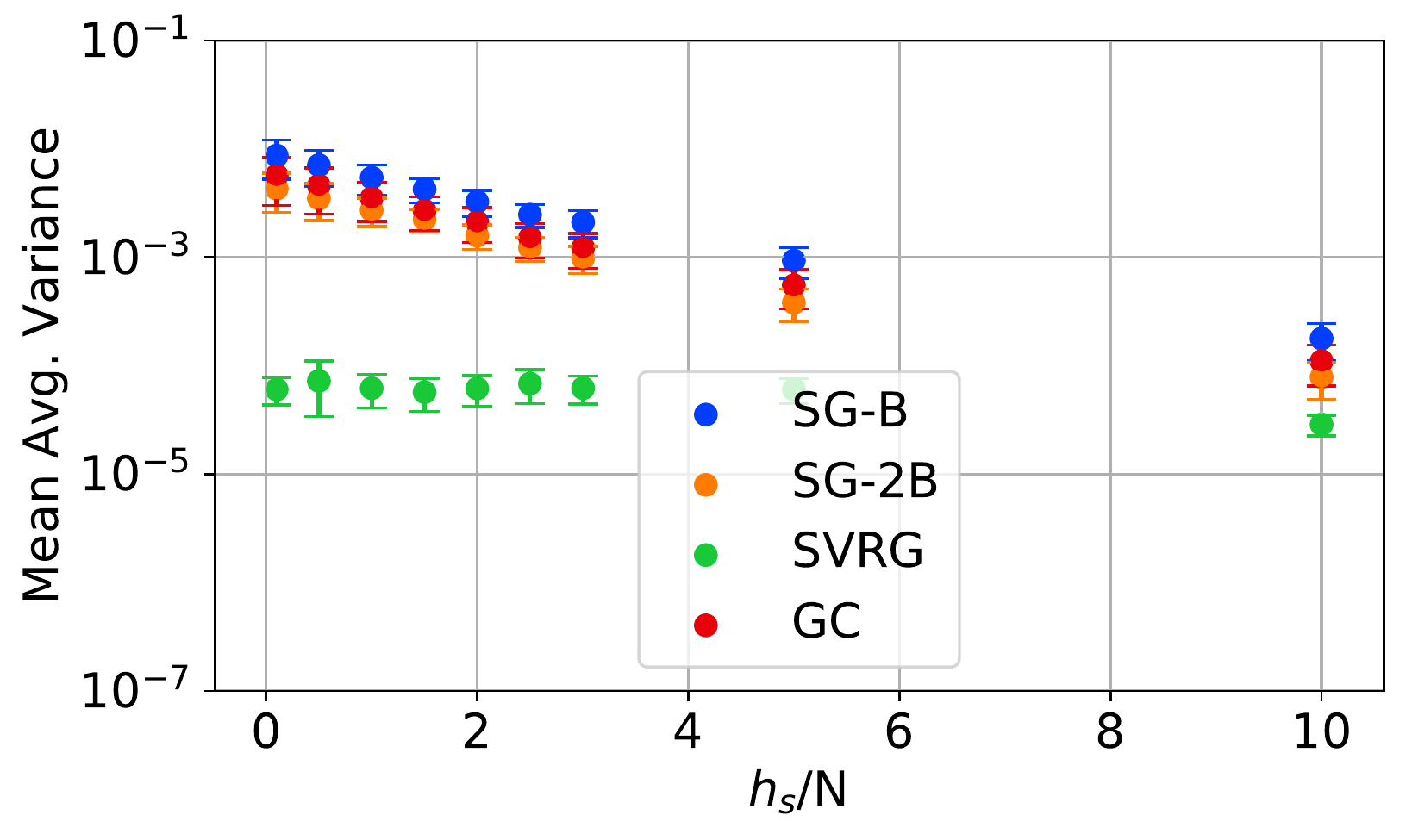}
    \hfill
    \includegraphics[width=.32\textwidth]{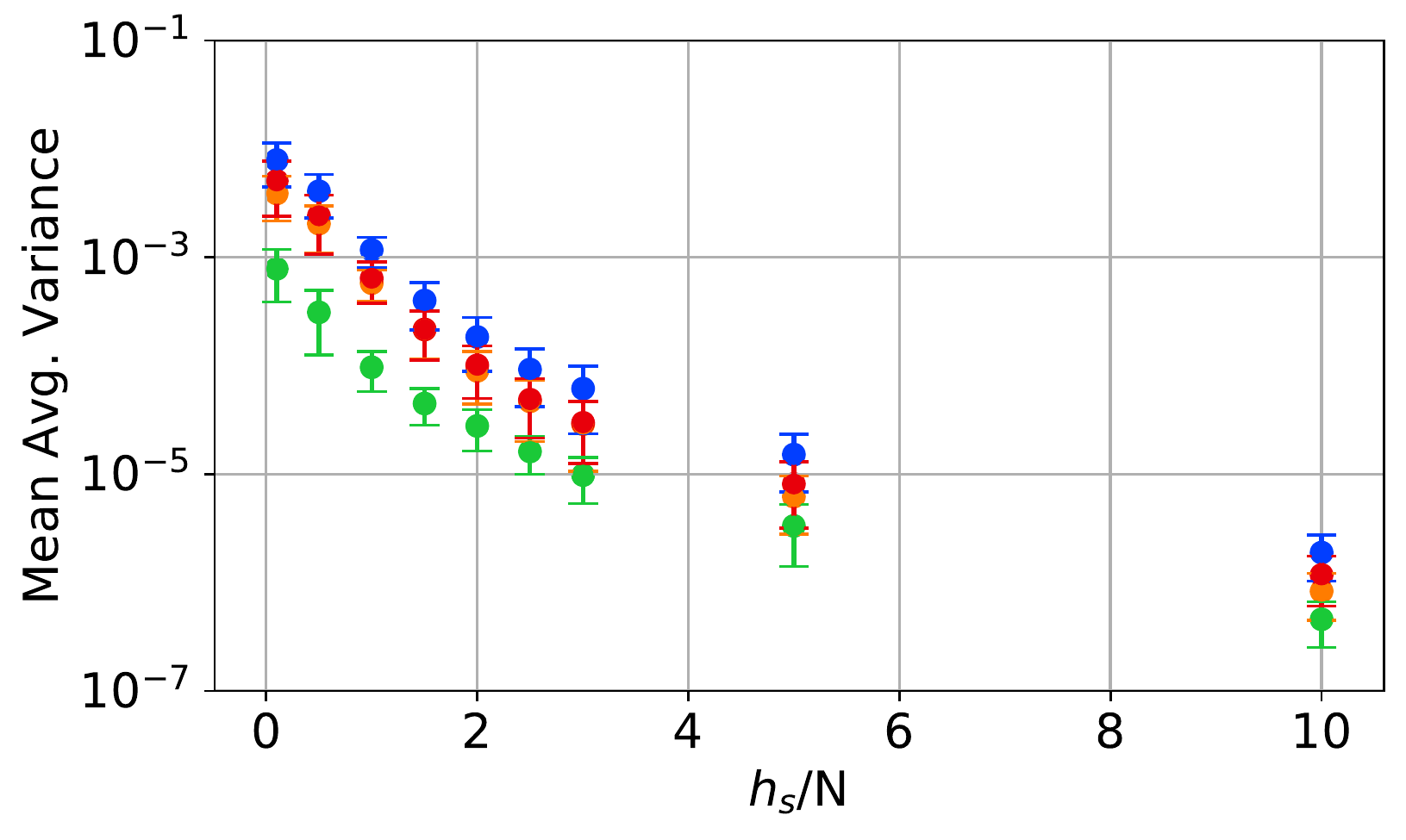}
    \caption{Mean variance plots for Fig.~\ref{fig:rf}}
    \label{fig:rf_mean}
\end{figure}

We aggregate results from multiple experiments with the following range of 
hyperparameters. Each point is generated by keeping $h_s$ fixed at $1000$ and 
varying $N$ in the range $[0.1, 10]$. We average over $3$ random seeds, $2$ 
teacher hidden dimensions and $2$ input dimensions (both $\times0.1$ and 
$\times10$ student hidden).

\end{document}